\documentclass[10pt,twocolumn,letterpaper]{article}

\usepackage{cvpr}
\usepackage{times}
\usepackage{epsfig}
\usepackage{graphicx}
\usepackage{amsmath}
\usepackage{amssymb}

\usepackage[ruled]{algorithm2e}
\usepackage{caption}
\usepackage{multirow}
\usepackage{diagbox}
\usepackage{footnote}
\usepackage{footmisc}
\usepackage{array}
\usepackage{graphicx}
\usepackage{subcaption}
\usepackage{amssymb}

\usepackage[pagebackref=true,breaklinks=true,letterpaper=true,colorlinks,bookmarks=false]{hyperref}

\cvprfinalcopy 

\def\cvprPaperID{633} 

\ifcvprfinal\fi
\begin{document}

\title{Location-aware Upsampling for Semantic Segmentation}

\author{Xiangyu He\and Zitao Mo\and Qiang Chen\and Anda Cheng\and Peisong Wang\and Jian Cheng\\
NLPR, Institute of Automation, Chinese Academy of Sciences, Beijing, China\\
{\tt\small {xiangyu.he, zitao.mo, qiang.chen, anda.cheng, peisong.wang, jcheng@nlpr.ia.ac.cn}}
}

\maketitle

\begin{abstract}
Many successful learning targets such as minimizing dice loss and cross-entropy loss have enabled unprecedented breakthroughs in segmentation tasks. Beyond these semantic metrics, this paper aims to introduce location supervision into semantic segmentation. Based on this idea, we present a Location-aware Upsampling (LaU) that adaptively refines the interpolating coordinates with trainable offsets. Then, location-aware losses are established by encouraging pixels to move towards well-classified locations. An LaU is offset prediction coupled with interpolation, which is trained end-to-end to generate confidence score at each position from coarse to fine. Guided by location-aware losses, the new module can replace its plain counterpart (\textit{e.g.}, bilinear upsampling) in a plug-and-play manner to further boost the leading encoder-decoder approaches. Extensive experiments validate the consistent improvement over the state-of-the-art methods on benchmark datasets. Our code is available at \href{https://github.com/HolmesShuan/Location-aware-Upsampling-for-Semantic-Segmentation}{https://github.com/HolmesShuan/Location-aware-Upsampling-for-Semantic-Segmentation}.
\end{abstract}

\section{Introduction}
Recent advances in deep learning have empowered a wide range of applications \cite{KrizhevskySH12, RenHGS15, resnet, ToshevS14, DongLHT16}, including semantic segmentation \cite{ZhaoSQWJ17, YuK15, LongSD15, ChenPKMY18}. The pioneering Fully Convolutional Network (FCN) \cite{LongSD15} with deconvolutional operations \cite{RonnebergerFB15} dramatically surpasses classical methods relying on hand-crafted features. Following FCNs, a series of CNN-based methods further enhance the state-of-the-art by introducing dilated (atrous) convolutions \cite{YuK15, ASPP}, shortcut connections \cite{BadrinarayananK17} and CRFs  \cite{VemulapalliT0C16, BertasiusTYS17, LinSHR16}. Latest semantic segmentation methods \cite{EncNet, deeplabv3} exploit fully trainable encoder-decoder architectures with simple bilinear upsamplings to extract high-resolution predictions.

Although recent works have exhibited state-of-the-art performance, we suggest that the location information hidden in the label map has been overlooked. To introduce location prediction into segmentation models, we
revisit the idea of semantic segmentation: \textit{the task of assigning each pixel of a photograph to a semantic class label} \cite{LempitskyVZ11}, then decompose this target into two relatively simple sub-tasks: pixel-level localization and classification. Instead of following previous methods to simply predict class labels, we further propose to predict locations simultaneously, more precisely, to learn offsets with additional supervision. 

This fresh viewpoint leads to the Location-aware Upsampling (LaU) directly, which upsamples classification results while accounting for the effect of offsets. Besides that, LaU tends to meet the key challenges in semantic segmentation: First, CNN-based segmentation methods extract high-level semantic features at the cost of losing detailed pixel-level spatial information. Second, the feature distribution of objects is imbalanced, where the interior points enjoy the rich knowledge after high-ratio upsampling yet pixels around edges suffer from indiscriminative features due to bilinear interpolation. A natural solution is to borrow the information from interior points to guide the categorization near the boundary, which exactly corresponds to our two sub-tasks: learning offsets and utilizing the information at the new position to make a prediction.

Our observation is that feature maps used by pixel-level classification, \textit{e.g.}, the second last layer in decoders, can also be used for predicting locations. On top of these features, we construct LaUs by adding an offset prediction branch to produce offsets $(\Delta x,\Delta y)$ at each position in the high-resolution feature map (details in section 3.3). To facilitate the offset prediction of interior points and improve the convergence, we first define the candidate locations $(\frac{x}{k},\frac{y}{k})$ then use predicted offsets to regress fine-grained coordinates, \textit{i.e.}, $(\frac{x}{k}+\Delta x,\frac{y}{k}+\Delta y)$. 

The proposed offset prediction makes it possible to further introduce location supervision implicitly/explicitly into the training process. In this paper, we present two location-aware losses forcing the offsets to predict fine-grained coordinates with the correct class. Besides, our location-aware losses avoid human labeling and produce ``label" online. In the experiments, we empirically show that location-aware mechanism outperforms strong baseline methods, and it is feasible to apply the rich knowledge in the design of location regression to the location-aware upsampling. We believe dense predictions can widely benefit from the location-aware module, including but not limited to the semantic segmentation.

\begin{figure}[t]
  \centering
  \includegraphics[width=3.3in]{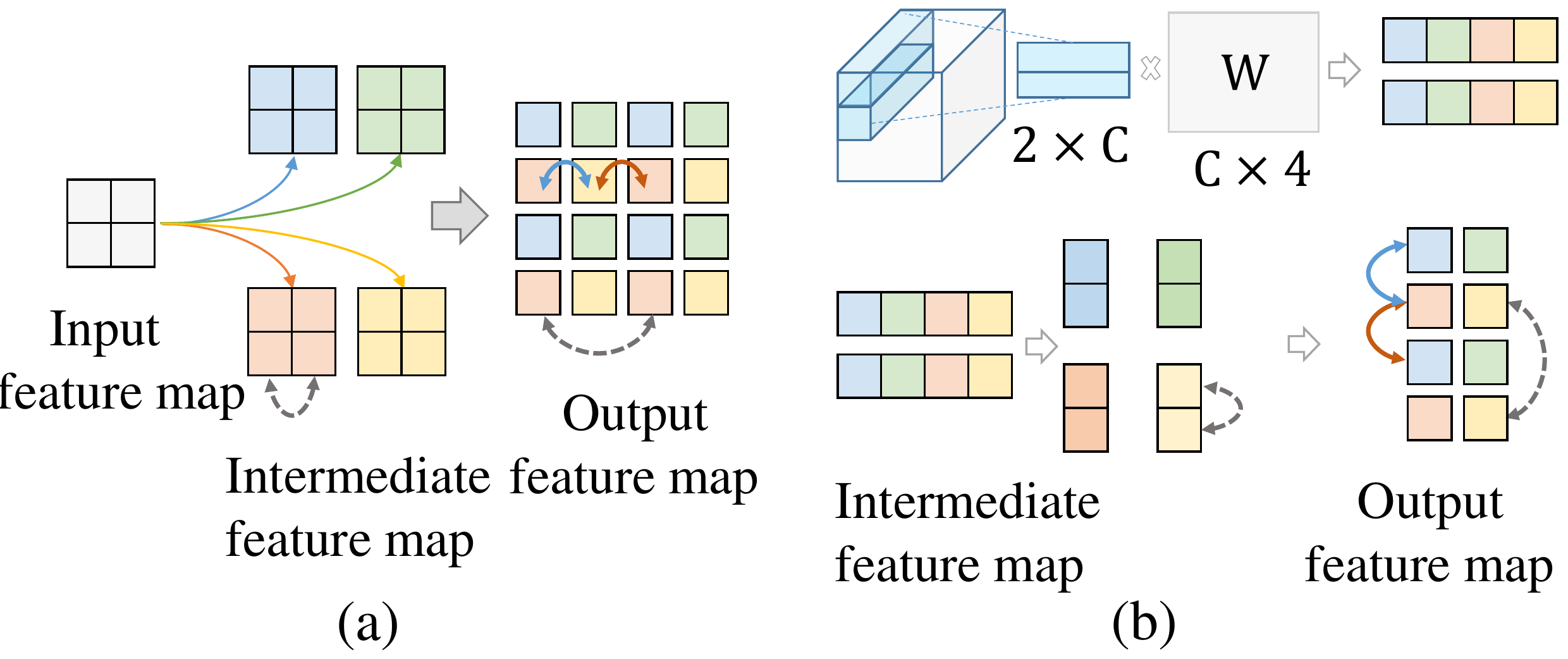}
\caption{(a) PixelShuffle \cite{ShiCHTABRW16} is in the form of periodical shuffling of multiple intermediate feature maps generated by independent filters, which cuts off the strong spatial relationship among adjacent pixels on the low-resolution feature map (connected by gray dashed lines). (b) DUpsampling \cite{DUpsampling} replaces convolutions with a matrix product to produce a small patch, yet can be transformed into PixelShuffle. In both cases, networks are forced to learn the new spatial relationship generated by hand-crafted rearrangement.}
\label{pixelshuffle}
\end{figure}

\section{Related Work}
Feature map upsampling has been widely used in deep models, such as encoder-decoder architectures \cite{NohHH15, RonnebergerFB15}, generative models \cite{RadfordMC15, MakhzaniF15}, super-resolution tasks \cite{DongLT16, ShiCHTABRW16, HeMWLY019} and semantic segmentation \cite{LongSD15, ChenPKMY18}. Deconvolution operations \cite{ZeilerKTF10, ZeilerTF11}, also known as transposed convolutions \cite{VedaldiL15}, first bridge the gap between low-resolution feature maps and desired full-resolution predictions. Despite its successes, recent works \cite{odena2016deconvolution, pixeldeconv} show that deconvolutions can easily suffer from the checkboard problem when the kernel size is not divisible by the stride. 

Advances like PixelShuffle \cite{ShiCHTABRW16} and DUpsampling \cite{DUpsampling} turn to hand-crafted pixel rearrangement to reconstruct high-resolution features. However, the empirical design results in the inconsistent spatial relationship among adjacent pixels on the high-resolution feature map (As illustrated in Figure \ref{pixelshuffle}, pixels connected by blue lines are generated by different filters, yet the input of each filter corresponds to the same receptive field. Both filter and receptive filed are different for pixels connected by red lines.). Currently, PixelDCL \cite{pixeldeconv} generates intermediate feature maps sequentially so that later stages depend on previously generated ones, which is similar to pixel-recurrent neural networks (PixelRNNs) \cite{OordKK16} and PixelCNNs \cite{OordKEKVG16}. Yet pixel-wise prediction is time-consuming and there is no explicit time sequence relation between adjacent pixels, \textit{i.e.}, previously generated pixels can still rely on later produced ones which leads to a chicken-and-egg problem.

Conditional Random Field (CRF) is another group of methods, that is primarily used in \cite{ChenPKMY14} as a disjoint post-processing. Following works \cite{LinSHR16, VemulapalliT0C16, LiuLLLT15} further integrate CRF into networks to model the semantic relationships between objects. Due to the increase of computing complexity caused by integration, \cite{ChandraK16, ChandraUK17} propose a Gaussian conditional random field to avoid the iterative optimization \cite{0001JRVSDHT15} while suffering from heavy gradient computation. Similar to CRFs, \cite{BertasiusTYS17, JiangGWTC18} focus on refining the output label map by random walks. However, the dense matrix inversion term and the two branch model with cascaded random walks cannot be easily merged into leading encoder-decoder architectures.

Spatial Transformer Networks (STN) \cite{JaderbergSZK15} is the first work to introduce spatial transformation capacity into neural networks, which enables global parametric transformation like 2D affine transformation. Since the pointwise location sampling in STN is weakly coupled with the input feature map at each position (only through global transformation parameters produced by the preceding feature maps), this greatly limits its capacity in pixel-wise predictions such as semantic segmentation. The module we present in this paper can be seen as a new transformation in a local and dense manner. The upsampled feature map can still enjoy the detailed information of low-resolution inputs by using pixel-wise offsets.

\begin{figure}[t]
  \centering
  \includegraphics[width=3.in]{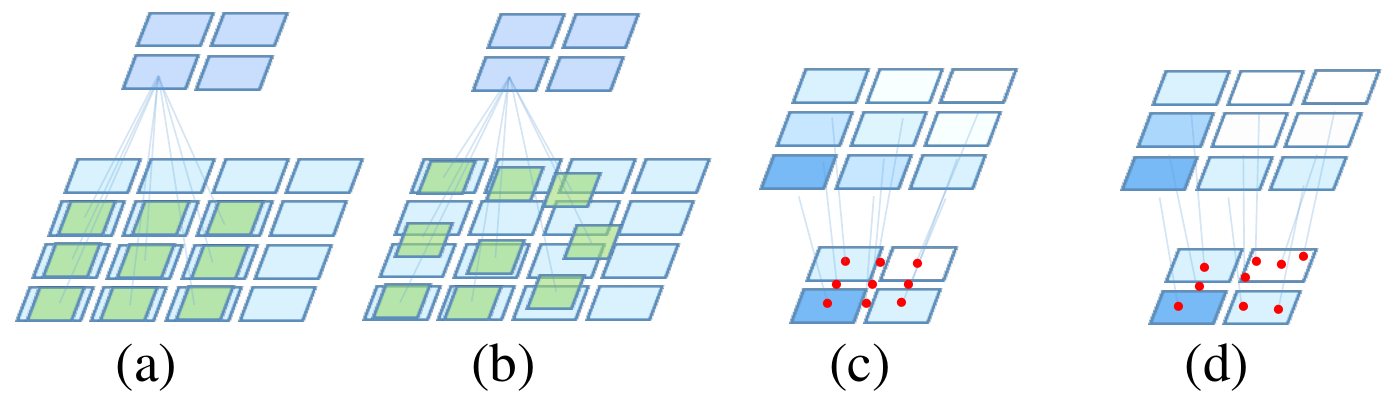} 
\caption{The illustrations of the original convolution (a), deformable convolution \cite{DaiQXLZHW17} (b), bilinear upsampling (c) and LaU (d).}
\label{DCNN}
\end{figure}

Deformable convolution networks (DCNN) \cite{DaiQXLZHW17, ZhuHLD19} achieve strong performance on object detection and semantic segmentation with trainable offsets. However, it is commonly used in feature extraction and we empirically prove that replacing the last bilinear upsampling with ``DCNN+Bilinear" will cause damage to the performance (details in Table \ref{deformable} and appendix section 2.2). As illustrated in Figure \ref{DCNN}, DCNN aims at augmenting the spatial sampling grid of filters  without extra supervision, yet our method directly refines the sampling coordinates with location-guided losses which contribute to higher performance.

\section{Methodology}
\begin{figure}
  \centering
  \includegraphics[width=3.3in]{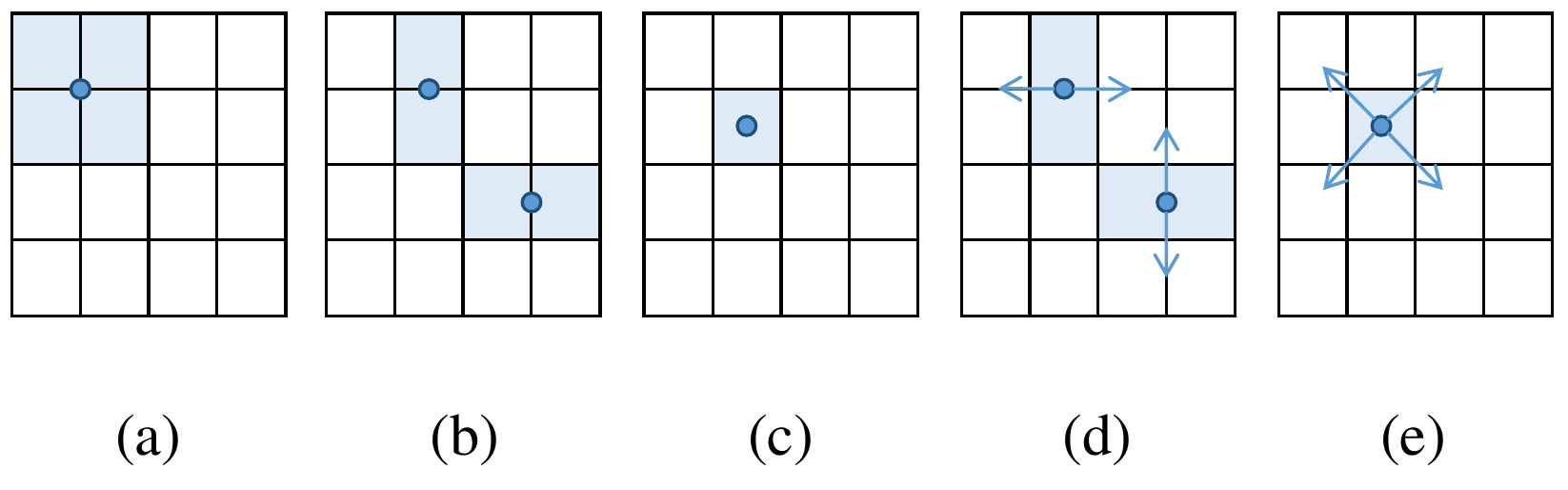} 
  \caption{Different cases in 2D bilinear interpolation. (a) An ideal case that fully exploits four points to obtain the desired estimate. (b) If one of the variables (\textit{i.e.}, $x$ or $y$) happens to be an integer, bilinear interpolation degrades into a linear interpolation.  (c) The special case :  both $x$ and $y$ are integers. Bilinear interpolation is an injective non-surjective function. (d)/(e) A natural solution to the degradation in (b,c) is to slightly move the interpolation point, which leads to another problem: how to determine the offsets? }  
\label{bilinear}
\end{figure}

Data-independent upsampling such as bilinear interpolation is one of the basic techniques in computer vision and image processing. Despite its high efficiency, this module can be oversimple (illustrated in Figure \ref{bilinear}) and fail to capture the semantic information. In this section, we alleviate the existing problems in bilinear upsampling by introducing the location-aware mechanism.

Learning data-dependent offsets is the key step in LaU. This idea is based on the fact that adjacent boundary points on the high-resolution label map can degrade into the same pixel in the low-resolution feature map, since $8\times$ upsampling is commonly used in popular methods \cite{ASPP,PSPNet,EncNet}. Instead of sampling from the local area, we propose to use the predicted offsets, generating the output map from the location-refined coordinates. We describe the formulation of a location-aware upsampling in the following sections.

\subsection{Spatial Upsampling}
To estimate the values in the output feature map $V\in\mathbb{R}^{C\times H\times W}$, a sampler applys the sampling kernel to a particular grid $\Omega$ in the input feature map $U\in\mathbb{R}^{C\times h\times w}$. Each coordinate $(x_j^u,y_j^u)$ defines the spatial location in $U$ and $\psi(\cdot)$ determines the kernel function at a particular pixel $V_i^{c_{\text{out}}}$ in channel $c_{\text{out}}$.  Formally, 
\begin{align}
V_i^{c_{\text{out}}}=\sum_{j}^{hw}\psi(x_j^u;x_i^v;\Phi_x)\text{ }\psi(y_j^u;y_i^v;\Phi_y)\text{ }U_j^{c_{\text{in}}}
\label{eq:1}
\end{align} 
where $\forall i\in[1,...,HW],\forall c_{\text{out}}\in[1,...,C]$, $(x_i^v,y_i^v)$ are the integer coordinates in the output feature map, $\Phi_x$ and $\Phi_y$ are the parameters of a generic sampling function $\psi(\cdot)$ which can be preset or data-dependent \footnote{$\Phi$ can be $0,1$ in $\text{max}(0,1-|x_j^u-x_i^v|)$, and $\Delta x_i^v,\Delta y_i^v$ in LaU.}. Note that the existing upsamplings such as bilinear and PixelShuffle apply the same $\psi(\cdot)$ to every channel of the input (\textit{i.e.}, data-independent), while LaU enables \textit{free changes of $\Phi$ at each position}. That is, the sampling kernel in LaU changes with the input data and coordinates.

In theory, $\psi(\cdot)$ is a (sub-)differentiable function of the coordinates $(x_i^v,y_i^v)$ and $(x_j^u,y_j^u)$, along with the parameters $\Phi$, which can be in any form. For example, $\psi(x_j^u;x_i^v;\Phi_x)=\text{max}(0,1-|x_j^u-x_i^v|)$ corresponds to the bilinear interpolation. Here, we show that PixelShuffle is another data-independent case, which may lead to sub-optimal results.

\subsubsection{PixelShuffle}
PixelShuffle can also be reformulated in the form of Eq.(\ref{eq:1}). Specifically, $\psi(\cdot)$ in PixelShuffle $2\times$ upsampling is
\begin{equation*}
\begin{aligned}
&\psi(x_j^u;x_i^v;\Phi_x)_{2\times}=\\
&\begin{cases}
\delta(x_i^v-2x_j^u) & \forall x_i^v\in[0,2,...,W\text{-2}], c_{\text{in}}\in[1,3,...,C\text{-1}] \\
\delta(x_i^v-2x_j^u-1) & \forall x_i^v\in[1,3,...,W\text{-1}], c_{\text{in}}\in[2,4,...,C]
\end{cases}
\end{aligned}
\end{equation*}
where $\delta(\cdot)$ is the Kronecker delta function. Note that $\Phi_x$ changes along with channel $c_{\text{in}}$, yet still fixed with respect to input data. From the view of $\Phi$, this sampling kernel equates to the data-independent methods like bilinear interpolation. Furthermore, the output of Kronecker delta is either $+1$ or $0$, which partly pinpoints the root of missing spatial relationship. That is, $\psi(\cdot)$ generated by PixelShuffle can be considered as a bijection. There is no direct relationship between adjacent elements in its codomain (output feature map), unless the source elements in its domain (intput feature map) are strongly correlated. For example, neighboring pixels connected by red lines in Figure \ref{pixelshuffle} belong to different receptive fields, and are generated by different filters; They are likely to look different (\textit{i.e.}, uncorrelated) unless the filter weights are similar/same. Therefore, the network has to consider the high-resolution spatial relationships when producing low-resolution intermediate feature maps.

\begin{figure*}
  \centering
  \includegraphics[width=6.4in]{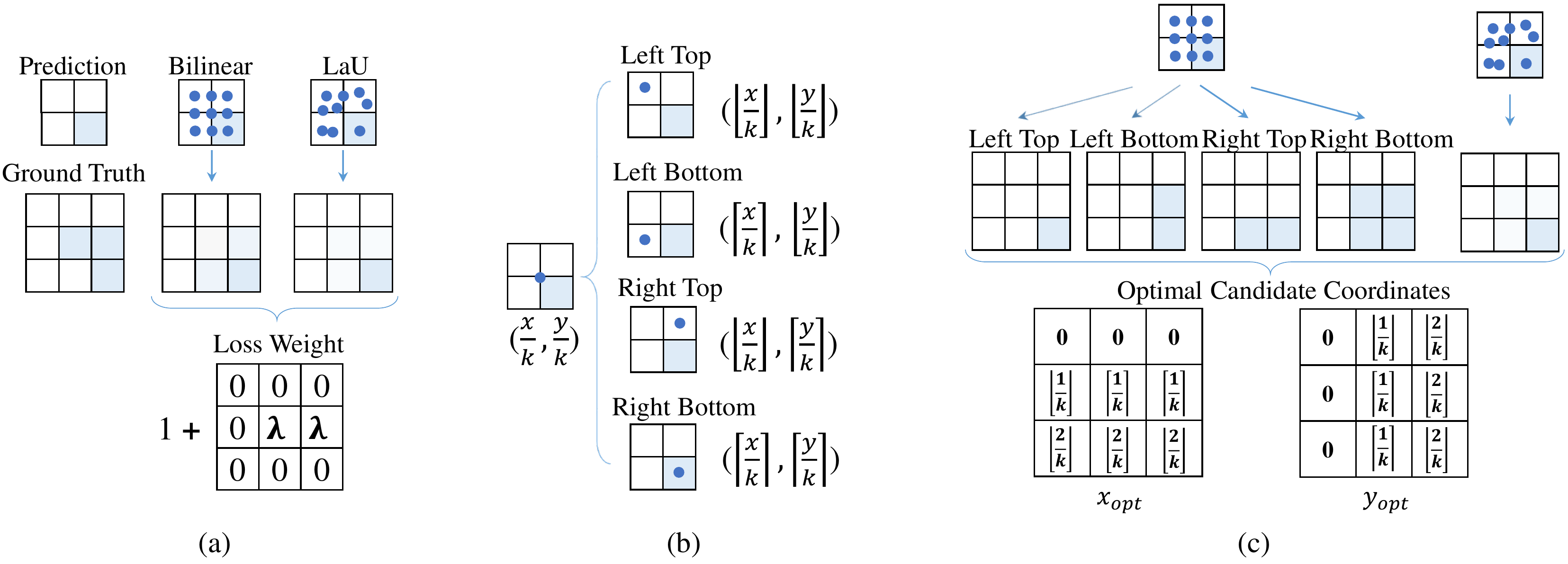} 
  \caption{Illustrations of the location-aware loss. $\downarrow$,$\searrow$,$\swarrow$ refer to the interpolation and upsampling. $\bullet$ is the coordinate point for interpolating. $\lceil\cdot\rceil$,$\lfloor\cdot\rfloor$ rounds input upward and downward, respectively. (a) Offset-guided loss will punish the moved pixel with loss weight $\lambda$ if it yields even larger loss than the original point. (b) The illustrations of different sampling kernels $\psi_{\text{left top}}$, $\psi_{\text{left bottom}}$, $\psi_{\text{right top}}$ and $\psi_{\text{right bottom}}$. (c) The optimal candidate coordinates correspond to pixels with the smallest loss among LaU, left top, left bottom, right top and right bottom upsamplings. Ground-truth is the same as subfigure (a).}
\label{loss}
\end{figure*}

\subsubsection{Location-aware Upsampling}
Mathematically, bilinear upsampling is a non-injective surjective function (not a bijection), denoted as $f:U\mapsto V$. Neighboring points in $V$ sampled from the same grid in $U$ are strongly correlated, since they are generated from the same elements. Compared with PixelShuffle, this scheme avoids checkerboard artifacts. As illustrated in Figure \ref{DCNN}, location-aware upsampling is a simple yet effective improvement to the bilinear interpolation with the following sampling kernel
\begin{align}
\psi(x_j^u;x_i^v;\Phi_x)&=\text{max}(0,1-|x_j^u-(\frac{x_i^v}{k}+\Delta x_i^v)|)
\label{lau}
\end{align}
where $k$ is the upsampling ratio. Inspired by the anchor-based methods in object detection \cite{Girshick15, RenHGS15}, we introduce $\frac{x_i^v}{k}$ as a candidate location then regress the detailed location with predicted offsets. To encourage the network to jump out of the local grid, we add no constraint on $(\Delta x_i^v, \Delta y_i^v)$. 

To learn the offsets in the upsampling module, we utlize the gradients with respect to $U$ and $(\Delta x_i^v,\Delta y_i^v)$. This allows the backpropagation of the loss function through the (sub-)differentiable location-aware upsampling module. For upsampling kernel (\ref{lau}), the partial derivatives are
\begin{align*}
&\begin{aligned}
\frac{\partial V_i^{c_{\text{out}}}}{\partial U_j^{c_{\text{in}}}}=\sum_j^{hw}\text{max}&(0,1-|x_j^u-(\frac{x_i^v}{k}+\Delta x_i^v)|)\cdot\\&\text{ max}(0,1-|y_j^u-(\frac{y_i^v}{k}+\Delta y_i^v)|)
\end{aligned}\\
&\begin{aligned}
\frac{\partial V_i^{c_{\text{out}}}}{\partial \Delta x_i^v}=\sum_j^{hw}U_j^{c_{\text{in}}}&\text{ max}(0,1-|y_j^u-(\frac{y_i^v}{k}+\Delta y_i^v)|)\cdot\\
&\begin{cases}
0 &\text{if  } |x_j^u-(\frac{x_i^v}{k}+\Delta x_i^v)|\geq1\\
-1 &\text{if  } \Delta x_i^v< x_j^u-\frac{x_i^v}{k}  \\
1 &\text{if  } \Delta x_i^v\geq x_j^u-\frac{x_i^v}{k}
\end{cases}.
\label{eq:5}
\end{aligned}
\end{align*}
With the gradients to offsets, we can further define the offset prediction branch to allow the end-to-end training.

\subsection{Location-aware Loss}
To fully exploit the effect of predicted offsets $(\Delta x_i^v,\Delta y_i^v)$, we introduce (semi-)supervised location-aware losses into the training process. Since offsets could point to any input location, as long as it contains the correct category, we first consider the offset-guided loss without hard constraints. Inspired by the bounding box regression, we further propose the offset regression loss to guide the offset prediction directly. 

\subsubsection{Offset-guided Loss}
In common training process of semantic segmentation, the network is optimized by per-pixel cross-entropy loss $\ell(\cdot)$. Instead of learning from isolated per-pixel supervision, we consider the  loss \textit{pair} generated by bilinear upsmapling and LaU (shown in Figure \ref{loss}a). Formally, we introduce the auxiliary loss $L'$ (with no gradient) and the target loss $L$,
\begin{equation*}
L=\ell(x,w,\psi_{\text{LaU}}(\Phi_x,\Phi_y)),\quad L'=\ell(x,w,\psi_{\text{bilinear}}) 
\end{equation*}
where $x$ is the input image, $w$ is the model parameters and $\Phi$ refers to the predicted offsets. The per-pixel offset-guided loss
\begin{equation}
\begin{aligned}
L^{\text{off}}_i=L_i\Lambda_i=L_i\cdot
&\begin{cases}
1 &\text{if  } L_i<L_i'\\
1+\lambda &\text{otherwise}
\end{cases}
\end{aligned}
\label{offsetloss}
\end{equation}
with loss weight $\Lambda_i$ forces the network to ``take a step", since the trivial solution $\Delta x,\Delta y=0$ (\textit{i.e.}, $L_i=L_i'$) will be punished by the extra cost $\lambda$. Besides, $L_i<L_i'$ demostrates that the moved pixel should have a smaller cross-entropy loss (\textit{i.e.}, a higher confidence score) than the original point. Therefore,  (\ref{offsetloss}) encourages the pixel to move towards a better position, which contains potentially correct class.  

\subsubsection{Regression Loss}
Since there is no groud-truth label of the offsets, we could intuitively search for the closest well classified point among the neighbouring  pixels to generate the offsets. However, the brute search is time-consuming for online label generation. Hence, we emprically constrain the search space to the intergral coordinate points (shown in Figure \ref{loss}b). 

Following the preivous subsection, we further introduce the auxiliary losses $L^{\text{lt}}$,  $L^{\text{lb}}$, $L^{\text{rt}}$ and $L^{\text{rb}}$, \textit{e.g.}, $L^{\text{rt}}=\ell(x,w,\psi_{\text{right top}})$ with the sampling kernels
\begin{equation*}
\psi_{\text{right top}}^x=1\{x_j^u=\left\lfloor\frac{x_i^v}{k}\right\rfloor\},\quad\psi_{\text{right top}}^y=1\{y_j^u=\left\lceil\frac{y_i^v}{k}\right\rceil\}
\end{equation*}
where $1\{\cdot\}$ is the ``indicator function", that is $1\{\text{true condition}\}=1$ and $1\{\text{false condition}\}=0$. Then, we define the loss set
\begin{align*}
\Omega=\{L,L^{\text{lt}}, L^{\text{lb}},L^{\text{rt}},L^{\text{rb}}\}
\end{align*}
and the coordinate set
\begin{align*}
\Theta=\Big\{(\frac{x}{k}+&\Delta x,\frac{y}{k}+\Delta y),~(\left \lfloor\frac{x}{k}\right \rfloor,\left\lfloor\frac{y}{k}\right\rfloor),\\&(\left\lceil\frac{x}{k}\right\rceil,\left\lfloor\frac{y}{k}\right\rfloor),
~(\left\lfloor\frac{x}{k}\right\rfloor,\left\lceil\frac{y}{k}\right\rceil),~ (\left\lceil\frac{x}{k}\right\rceil,\left\lceil\frac{y}{k}\right\rceil)\Big\}.
\end{align*}
Combining $\Omega$ and $\Theta$, we obtain the optimal candidate corrdinates
\begin{align*}
\Theta^{\text{opt}}=\sum_{k}1\{\text{min}(\Omega)=\Omega_k\}\cdot\Theta_k
\end{align*}
where the summation stops at $1\{\text{min}(\Omega)=\Omega_k\}=1$. Given $\Theta^{\text{opt}}$ (illustrated in Figure \ref{loss}c), we formulate the per-pixel offset regression loss as
\begin{align}
\begin{aligned}
L^{\text{reg}}_i&=\gamma\text{SmoothL}_1\Big(\Theta^{\text{opt}}_i,(\frac{x_i^v}{k}+\Delta x_i^v,\frac{y_i^v}{k}+\Delta y_i^v)\Big)+L_i\Lambda_i\\
&=\gamma\text{SmoothL}_1+L_i\cdot
\begin{cases}
1 &\text{if  } L_i\leq\forall\Omega_i\\
1+\lambda &\text{otherwise}
\end{cases}.
\end{aligned}
\end{align}
where $\gamma$ controls the strength of the regression loss. In our experiments, we set $\gamma$ to 0.1 without fine-tuning. Note that the auxiliary losses will be excluded from the backward propagation (with no gradient) and omitted at inference time. 

\subsection{Location-aware Network}
\begin{figure}
  \centering
  \includegraphics[width=2.7in]{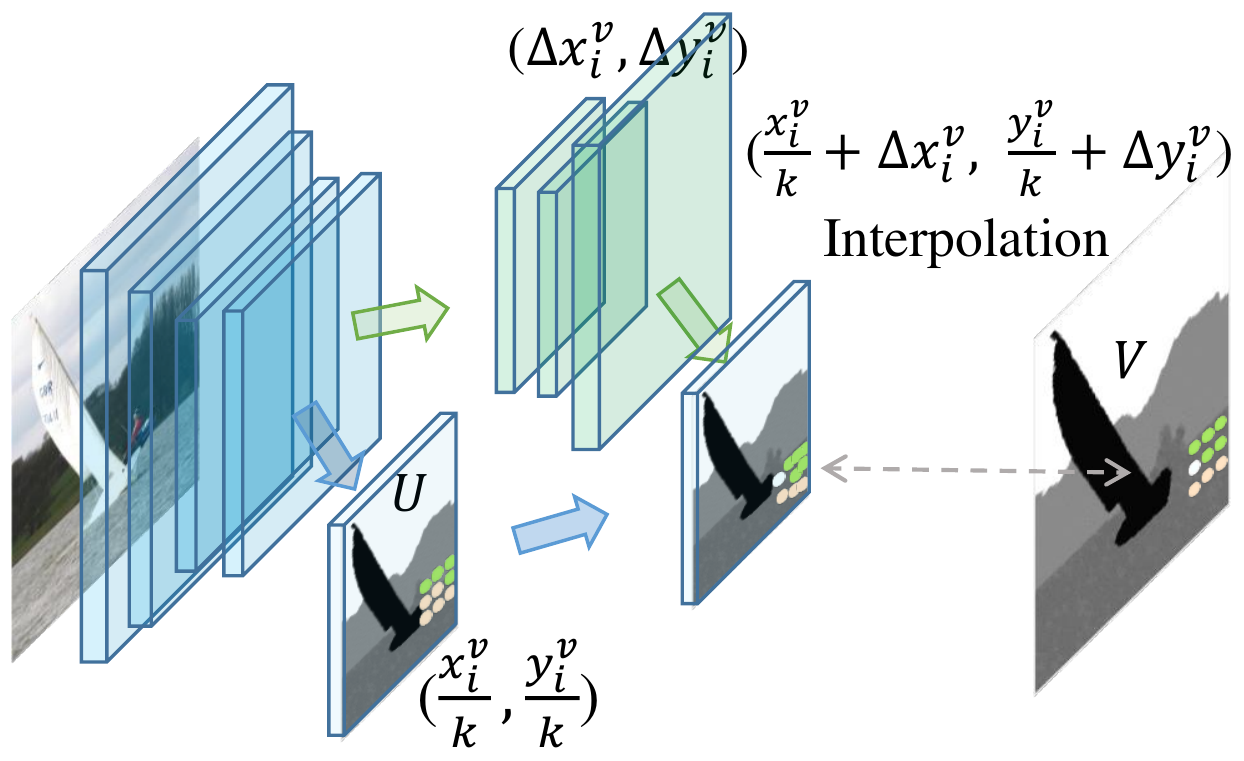} 
  \caption{The framework overview of LaU. Our approach employs the same encoder-decoder architecture as the original method, which adds a $1\times1$ convolution layer over the decoder output to generate per-pixel classification scores (\textit{i.e.}, $U\in\mathbb{R}^{C\times h\times w}$). The offset prediction branch (green blocks) shares the input feature map with the $1\times1$ convolution then produces $(\Delta x_i^v,\Delta y_i^v)$. A location-aware upsampling merges two branches and generates high-resolution outputs.  }
\label{LaU_framework}
\end{figure}

The combination of predicted offsets and interpolation operation forms a location-aware upsampling (Figure \ref{LaU_framework}). Note that the offset prediction branch can take any form in practice, such as a fully-connected network like RPN in Faster-RCNN \cite{RenHGS15}. In this paper, we use ``Conv$_{1\times1}$+LReLU+Conv$_{3\times3}$+PixelShuffle" to illustrate the basic framework of location-aware mechanism in semantic segmentation. 

The first $1\times1$ convolution layer reduces the channel number of the input feature map $U\in\mathbb{R}^{C\times h\times w}$ from $C$ to $C'$. Then, the last convolution layer produces an offset tensor $O\in\mathbb{R}^{2M\times H\times W}$ where $H=kh$ and $W=kw$. $M$ can be the number of class to allow the per-element shift. However, the extra computing cost will be unacceptable if $M$ is too large. When we set $M$ to 1, the output feature map shares offset across different channels. Compared with $M=150$, this setting saves over $19$G FLOPs on ADE20K dataset.

Encoder-decoder architecture with LaU tends to first predict the general location and rough boundary, then recontruct the detailed information using predicted offsets. As shown in Figure \ref{6c} and \ref{6d}, this two-step scheme decomposes the original problem into two sub-tasks: pixel-level localization and classification, which correspond to the two branches in location-aware network.

\section{Experiments}
\begin{figure}
\center
    \begin{subfigure}[t]{0.11\textwidth}
            \begin{subfigure}[t]{\textwidth}
                \includegraphics[width=\textwidth]{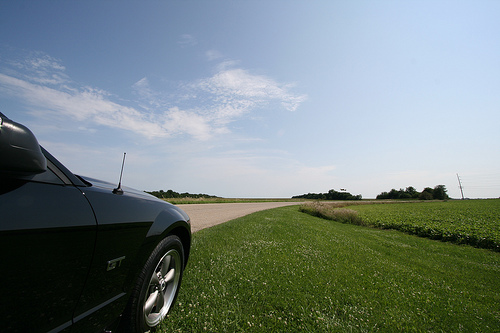}
            \end{subfigure}\vspace{.6ex}

            \begin{subfigure}[t]{\textwidth}
                \includegraphics[width=\textwidth]{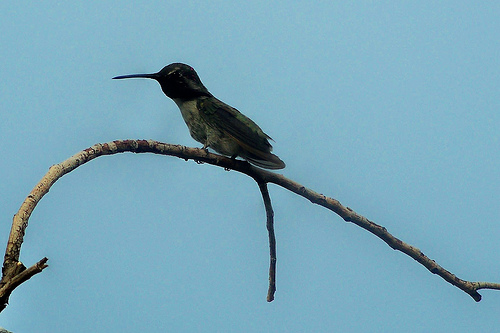}
            \end{subfigure}\vspace{.6ex}

         \begin{subfigure}[t]{\textwidth}
                \includegraphics[width=\textwidth]{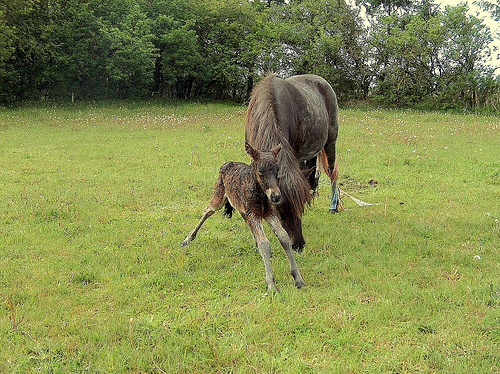}
            \end{subfigure}\vspace{.6ex}

            \begin{subfigure}[t]{\textwidth}
                \includegraphics[width=\textwidth]{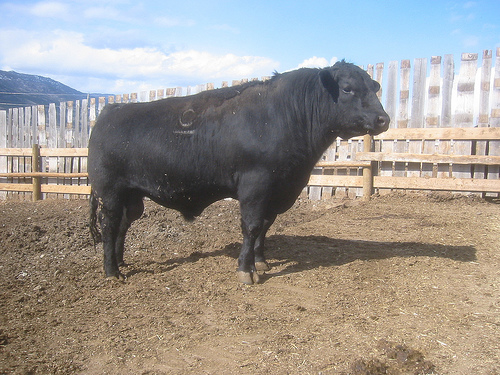}
            \subcaption{Image}
            \end{subfigure}
     \end{subfigure}
    \begin{subfigure}[t]{0.11\textwidth}
        \begin{subfigure}[t]{\textwidth}
                \includegraphics[width=\textwidth]{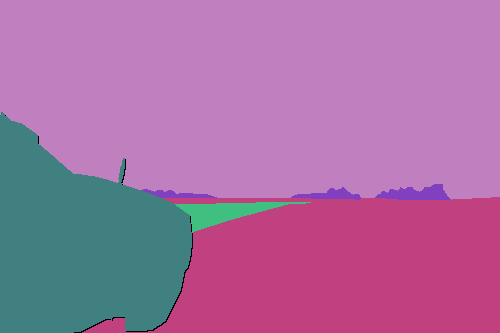}
            \end{subfigure}\vspace{.6ex}

        \begin{subfigure}[t]{\textwidth}
                \includegraphics[width=\textwidth]{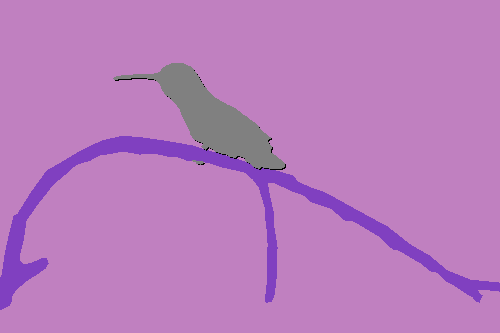}
            \end{subfigure}\vspace{.6ex}

        \begin{subfigure}[t]{\textwidth}
                \includegraphics[width=\textwidth]{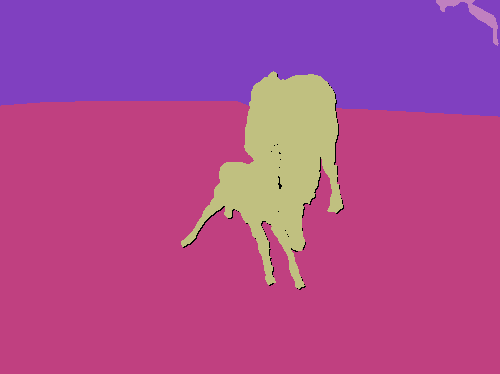}
            \end{subfigure}\vspace{.6ex}

            \begin{subfigure}[t]{\textwidth}
                \includegraphics[width=\textwidth]{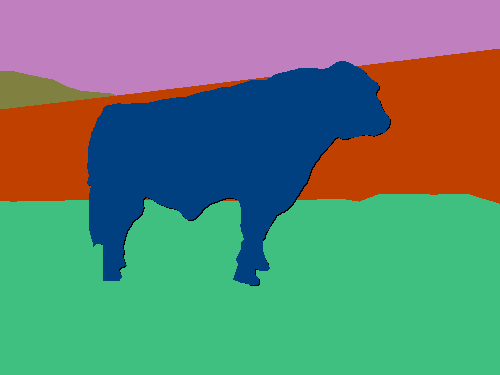}
     \captionsetup{justification=centering}       
	\subcaption{GT}
            \end{subfigure}
    \end{subfigure}
    \begin{subfigure}[t]{0.11\textwidth}
        \begin{subfigure}[t]{\textwidth}
                \includegraphics[width=\textwidth]{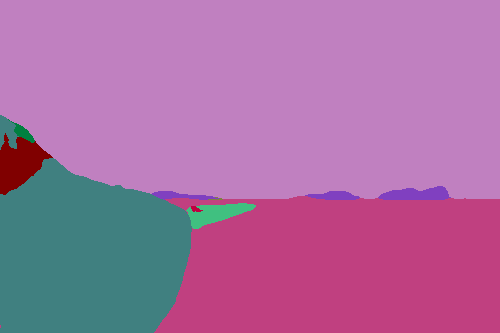}
            \end{subfigure}\vspace{.6ex}

        \begin{subfigure}[t]{\textwidth}
                \includegraphics[width=\textwidth]{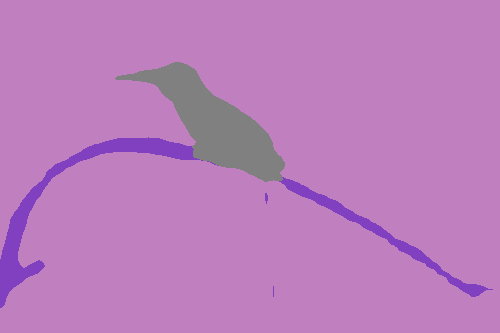}
            \end{subfigure}\vspace{.6ex}

        \begin{subfigure}[t]{\textwidth}
                \includegraphics[width=\textwidth]{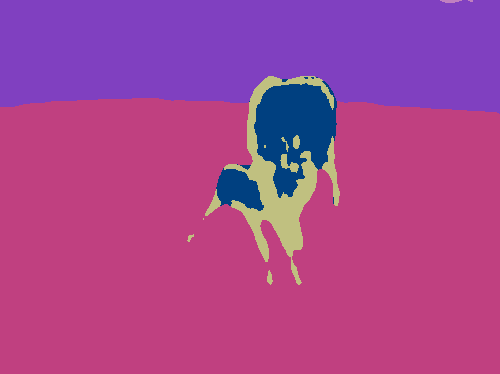}
            \end{subfigure}\vspace{.6ex}

            \begin{subfigure}[t]{\textwidth}
                \includegraphics[width=\textwidth]{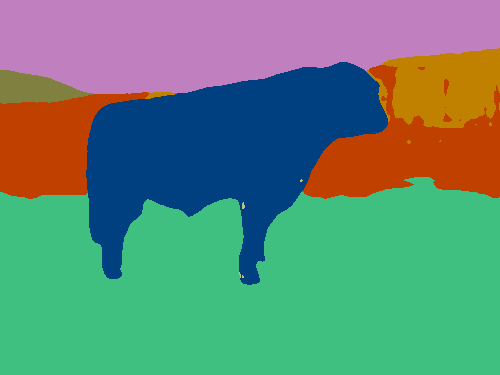}
            \captionsetup{justification=centering}
            \subcaption{EncNet}
        \label{7c}
            \end{subfigure}
    \end{subfigure}
    \begin{subfigure}[t]{0.11\textwidth}
        \begin{subfigure}[t]{\textwidth}
                \includegraphics[width=\textwidth]{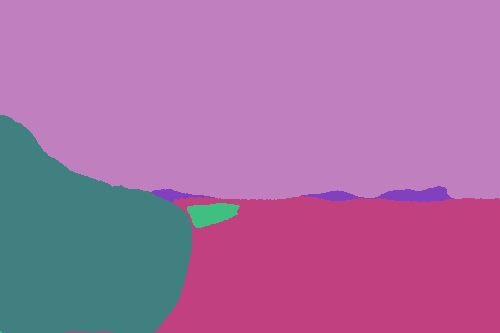}
            \end{subfigure}\vspace{.6ex}

        \begin{subfigure}[t]{\textwidth}
                \includegraphics[width=\textwidth]{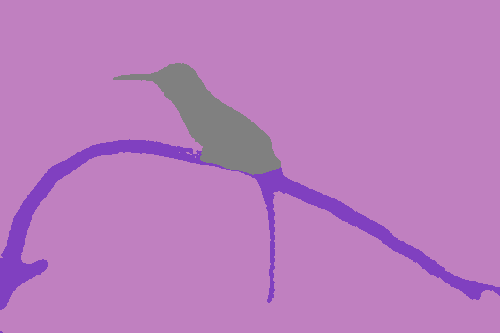}
            \end{subfigure}\vspace{.6ex}

        \begin{subfigure}[t]{\textwidth}
                \includegraphics[width=\textwidth]{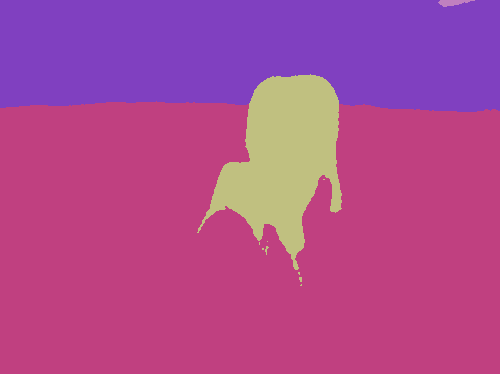}
            \end{subfigure}\vspace{.6ex}

            \begin{subfigure}[t]{\textwidth}
                \includegraphics[width=\textwidth]{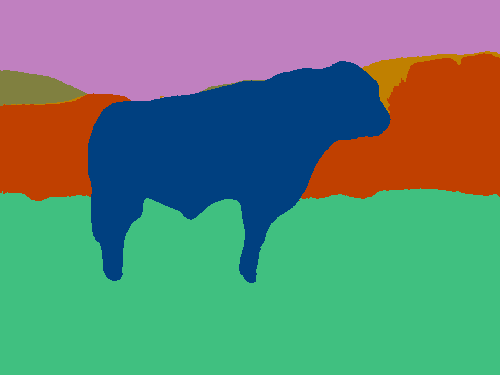}
            \subcaption{LaU}
        \label{7d}
            \end{subfigure}
    \end{subfigure}
        \caption{Visualization results of different upsampling modules using EncNet-ResNet50 \cite{EncNet}. "EncNet" refers to the original bilinear upsampling. "LaU" is $4\times$ location-aware upsampling with offset-guided loss.}
\label{view}
\end{figure}

We evaluate the proposed method on the PASCAL VOC 2012 semantic segmentation benchmark \cite{EveringhamGWWZ10}, PASCAL Context \cite{MottaghiCLCLFUY14} and ADE20K \cite{ADE20K} datasets. The standard evaluation metrics are pixel accuracy  (pixAcc) and mean Intersection of Union (mIoU). In this section, we first introduce datasets used in this paper as well as the implementation details. We then apply LaU to the top of networks and the upsampler in decoders to show the effectiveness of the location-aware mechanism. Moreover, the visualization results demonstrate the superiority of optimizing the proposed location-aware segmentation (shown in Figure \ref{view}). 

\subsection{Experimental Settings}
\paragraph{Dataset} PASCAL Context dataset \cite{MottaghiCLCLFUY14} provides pixel-wise semantic annotations for whole scenes, which contains 4998 images for training and 5105 images for testing. Following \cite{LinMSR17, DANet, JPU, EncNet}, we use the most frequent 59 object categories plus background (60 classes in total) as the groundtruth labels. PASCAL VOC 2012 consists of 1464 (\textit{train}), 1449 (\textit{val}) and 1456 (\textit{test}) pixel-level annotated images, which involves 20 foreground object classes and one background class. Following previous works \cite{deeplabv3, HariharanABMM11}, we augment the dataset by the extra annotations, resulting in 10582 (\textit{trainaug}) training images. The mIoU is averaged across 21 classes. ADE20K \cite{ADE20K} scene parsing benchmark consists of 150 categories. There are 20K images for training, 2K for validation and 3K for testing. The results on test set are provided by the ADE20K online evaluation server.

\paragraph{Implementation Details}
For the experiments on PASCAL Context, we adopt the same settings as described in \cite{EncNet, JPU}. The learning rate starts at 0.001, which gradually decrease to 0 using the "poly" strategy (power=0.9). The networks are trained for 80 epochs with SGD. Concretely, the momentum is set to 0.9 and weight decay is 0.0001. To encourage the LaU to "look wider", we set the weight decay of the convolutional layer in LaU to 0. For data augmentation, we randomly scale (from 0.5 to 2.0)  and horizontally flip the input images, then apply a $480\times480$ random crop to the images. All the experiments are conducted on 4 GPUs with batch size 16. Since ImageNet pre-trained backbones can significantly contribute to higher performances, we follow the practice of \cite{EncNet, JPU, DANet, DUpsampling} to use the standard ResNet-50 and ResNet-101 \cite{resnet}. 

For training on PASCAL VOC, we follow the setting in \cite{DUpsampling, deeplabv3}. The initial learning rate is set to 0.007 with "poly" policy and weight decay is 0.0001. The total iteration is $30k$ for ResNet-50 experiments with batch size 48, yet another $30k$ iterations are required for training ResNet-101 with initial learning rate being 0.001 \cite{deeplabv3}. The scaling with flipping data augmentation and LaU settings remain the same as PASCAL Context. 

For the experiments on ADE20K, as described in EncNet \cite{EncNet}, we train our model on the training set for 180 epochs with learning rate 0.004 and evaluate it on the validation set using multi-scale testing. We then fine-tune our models on the \textit{train-val} set for another 30 epochs with learning rate $4e^{-4}$ and submit results to the test server. We use a $520\times520$ base size with $480\times480$ crop for training. The backbone networks are the same as previous experiments.

Unless specified, we employ the same loss function as described in \cite{ASPP, EncNet} plus a location-aware loss and report the single-scale results, \textit{without any bells and whistles}.

\begin{table}[t]
  \caption{Performance on Pascal Context \textit{val} set with EncNet-ResNet50 (60 classes) using different upsampling ratios. ``LaU$_{\text{off}}$" refers to LaU with offset-guided loss. We cover the reamining upsampling factor of LaU with bilinear upsampling.}
  \label{sample-table}
  \centering
\begin{tabular}{lcccc}
\hline 
    Upsampling     & Ratio     & mIoU\% & Params & FLOPs\\
\hline   
\hline
 Bilinear (\cite{EncNet}) & $8\times$ & 49.08 & 41.82M  & 150.69G\\
\hline 
    \multirow{4}{*}{LaU$_{\text{off}}$} & $2\times$ & 49.61 & $+0.14$M & $+0.51$G\\
    & $3\times$ & 49.67 & $+0.15$M & $+0.55$G\\
    & $4\times$ & \textbf{50.27} & $+0.16$M & $+0.60$G\\
    & $5\times$ & 49.65 & $+0.18$M & $+0.68$G\\
\hline 
  \end{tabular}
\label{diffratio}
\end{table}

\begin{table}[t]
  \caption{Ablation study on loss weight $\lambda$ using EncNet-ResNet50+LaU$_{\text{off}_{4\times}}$ (Pascal Context \textit{val} 60 classes).}
  \label{sample-table}
  \centering
\begin{tabular}{cccccc}
\hline 
    $\lambda$   & 0.0   & 0.1 & 0.2 & 0.3 & 0.4\\
\hline   
\hline
 mIoU\% & 49.77  & 49.84 & 49.94 & \textbf{50.27} & 49.98 \\
pixAcc\% & 78.69 & 78.77 & 78.75 & \textbf{79.01} & 78.89 \\
\hline 
  \end{tabular}
\label{lambda}
\end{table}

\paragraph{Ablation Study for LaU}
To evaluate LaU, the key parameter is the upsampling factor. As listed in Table \ref{diffratio}, $4\times$ upsampling works better than other settings with negligible additional time and space complexity. With proposed LaU, the results consistently outperform standard bilinear upsampling and the best setting exceeds baseline by +1.2 mIoU with only 0.16M extra parameters. Since loss weight of the original auxiliary loss in EncNet is 0.2, in this section, we intuitively set $\lambda$ to 0.3 to enlarge the gradient of location-aware loss.

\begin{table*}[t]
    \caption{Semantic segmentation results on Pascal Context \textit{val} set. mIoU on 60 classes w/ background.}
    \begin{subtable}{.52\linewidth}
        \caption{Applying to encoder-decoder methods with ResNet-50. ``OS" refers to the output stride. We first apply an upsampling with LaU/PixelShuffle, then cover the remaining upsampling factor with bilinear upsampling. ``$2\times+2\times$" means the cascade of two ${2\times}$, \textit{i.e.}, $4\times$.}
      \newcommand{\tabincell}[2]{\begin{tabular}{@{}#1@{}}#2\end{tabular}}
        \begin{tabular}{p{2.cm}<{\centering}lcc}
\hline
            Head  & Upsampling & OS & mIoU\% \\
\hline
\hline
         \multirow{5}{*}{EncNet \cite{EncNet}} & PixelShuffle$_{2\times}$  & 8 & 43.4 \\
		& PixelShuffle$_{3\times}$  & 8 &  38.3 \\
           & PixelShuffle$_{2\times+2\times}$  & 8 & 27.9 \\
		& Bilinear  & 8 & 49.1\\
          & LaU$_{\text{off}_{4\times}}$ & 8 & \textbf{50.3} \\
\hline
\multirow{3}{*}{EncNet \cite{EncNet}} & PixelShuffle$_{2\times}$  & 32 & 37.8 \\
           & Bilinear  & 32 & 45.7 \\
           & LaU$_{\text{off}_{4\times}}$  & 32 & \textbf{46.7} \\
\hline
        \multirow{3}{*}{\tabincell{c}{ASPP \cite{ASPP}}}   & PixelShuffle$_{2\times}$  & 8 & 43.3 \\
		& Bilinear  & 8 & 48.4\\
          & LaU$_{\text{off}_{4\times}}$ & 8 & \textbf{49.8} \\
\hline
        \end{tabular}
    \label{pc_res50}
    \end{subtable}%
    \begin{subtable}{.55\linewidth}
      \centering
       \captionsetup{justification=centering}
        \caption{Results with state-of-the-art methods. ``off$_{4\times}$'' and ``reg$_{4\times}$'' are offset-guided loss and offset regression loss, respectively. \\"MS" means multi-scale testing.  }
        \begin{tabular}{llcc}
\hline
            Method  & Backbone & mIoU\% & MS \\
\hline
\hline
         CCL \cite{DingJSL018} & ResNet-101 & 50.7 & \\ 
         DUpsampling \cite{DUpsampling} & Xception-65 & 51.4 & $\checkmark$\\
         DUpsampling \cite{DUpsampling} & Xception-71 & 52.5 & $\checkmark$\\
         DANet \cite{DANet} & ResNet-101 & 52.6 & $\checkmark$\\
	    HRNet \cite{HRR} & HRNetV2-W48  & 53.1 & $\checkmark$\\
	    JPU \cite{JPU} & ResNet-101 & 53.1 & $\checkmark$\\
	    BFP \cite{BFP} & ResNet-101 & 52.7 & $\checkmark$\\
         EncNet \cite{EncNet} & ResNet-101 & 51.7 & $\checkmark$\\
\hline
         EncNet + LaU$_{\text{off}_{4\times}}$ & ResNet-101 &  \textbf{53.9} & $\checkmark$\\
	   EncNet + LaU$_{\text{off}_{4\times}}$ & ResNet-101 &  52.7 & \\
	   EncNet + LaU$_{\text{reg}_{4\times}}$ & ResNet-101 &  \textbf{53.9} & $\checkmark$\\ 
	   EncNet + LaU$_{\text{reg}_{4\times}}$ & ResNet-101 &  52.8 & \\
\hline
        \end{tabular}
    \label{pc_res101}
    \end{subtable} 
\end{table*}

\paragraph{Ablation Study for Location-aware Loss}
We experiment with loss weight $\lambda$ between 0 and 0.4 and show the results in Table \ref{lambda}. Note that LaU without additional supervision ($\lambda=0$) still produces  performance gain and $\lambda=0.3$ yields the best performance. Both location-aware upsampling and location-aware loss contribute to higher performance than bilinear baseline (similar results are reported in Table \ref{deformable}). Due to the limited computing resources, we directly apply the best settings to regression loss experiments and emprically set $\gamma$ to 0.1 without grid search.

\subsection{Applying to Label Prediction}
In this section, we apply LaU to the top of the decoders (as shown in Figure \ref{LaU_framework}), which corresponds to the prediction of label maps.
\subsubsection{Quantitative Analysis}
\paragraph{PASCAL Context} To estimate the generalization ability of LaU, we conduct experiments on two popular methods Encoding \cite{EncNet} and ASPP \cite{ASPP} with different output strides. Table \ref{pc_res50} shows that our methods consistently outperform the original bilinear methods. Following the setting in super-resolution tasks \cite{LinSHR16,HeMWLY019,ShiCHTABRW16}, one may decompose a single $8\times$ upsampling into three cascaded $2\times$ upsamplings. To simplify the experimental settings, we leave this scheme as future work.

Compared with PixelShuffle \cite{ShiCHTABRW16} (Table \ref{pc_res50}) and Deformable \cite{DaiQXLZHW17} bilinear upsampling (Table \ref{deformable}), LaU notably outperforms those methods with/without location-aware loss. Note that PixelShuffle and Deformable cause damages to the performance; We show the visualization results and give a brief discussion in the appendix section 2. 

Table \ref{pc_res101} further shows that single-scale LaUs surpass recent upsampling method DUpsampling \cite{DUpsampling} with strong Xception-71 backbone, \textit{e.g.}, ImageNet Top-1 accuracy of Xception-65 is $+2.21\%$ higher than ResNet-101. Our multi-scale result achieves $+2.2$ mIoU over EncNet, while LaU is still compatible with attention mechanisms and stronger encoder-decoder architectures to further enhance the leading performances.
\begin{table}
  \caption{Comparison with Deformable Convolution Netwrok \cite{DaiQXLZHW17} on Pascal Context \textit{val} set. We add an ``LaU/Deformable+Bilinear" module to the top of EncNet/ASPP. ``LaU$_{4\times}$" refers to LaU without location-aware loss. ``OS" is the output stride.}
  \centering
 \begin{tabular}{l|lcc}
\hline    
Method    & Upsampling & OS   & mIoU\%\\
\hline  \hline  
    \multirow{6}{*}{EncNet \cite{EncNet}}  & Deformable+Bilinear & 32  & 44.7 \\
    & LaU$_{4\times}$ & 32 & 46.2 \\
    & LaU$_{\text{off}_{4\times}}$ & 32 & \textbf{46.7} \\
\cline{2-4}  
& Deformable+Bilinear & 8  & 48.0  \\
& LaU$_{4\times}$ & 8 & 49.8 \\
    & LaU$_{\text{off}_{4\times}}$ & 8 & \textbf{50.3} \\
\hline
\multirow{3}{*}{ASPP \cite{ASPP}}  & Deformable+Bilinear & 8  & 47.9 \\
& LaU$_{4\times}$ & 8 & 49.4 \\
    & LaU$_{\text{off}_{4\times}}$ & 8 & \textbf{49.8} \\
\hline  
  \end{tabular}
\label{deformable}
\end{table}
\paragraph{ADE20K} As shown in Table \ref{ade20k_val}, our methods achieve similar results on ADE20K dataset. With ResNet-50 backbone, LaUs outperform EncNet baseline by $+1.8$ and $+1.3$ mIoU. Regression loss seems to perform worse than location-guided loss. We attribute this to the hand-crafted design of optimal candidate coordinates and the selection of $\gamma$. The current search space contains only four points, which can be suboptimal. We will include a larger search space in our future work.

Although our methods achieve slightly worse performance than EncNet on \textit{val} set, LaUs notably surpass EncNet and all entries in COCO-Place challenge 2017 on the final \textit{test} score (see results in Table \ref{ade20k_test}).

\subsubsection{Qualitative Analysis}
\begin{figure}[t]
\center
    \begin{subfigure}[t]{0.11\textwidth}
            \begin{subfigure}[t]{\textwidth}
                \includegraphics[width=\textwidth]{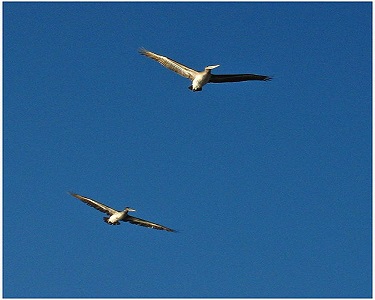}
            \end{subfigure}\vspace{.1ex}

            \begin{subfigure}[t]{\textwidth}
                \includegraphics[width=\textwidth]{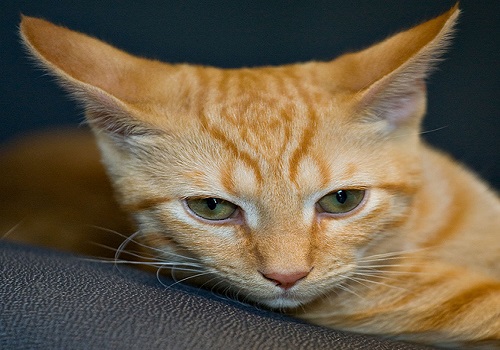}
            \end{subfigure}\vspace{.1ex}

         \begin{subfigure}[t]{\textwidth}
                \includegraphics[width=\textwidth]{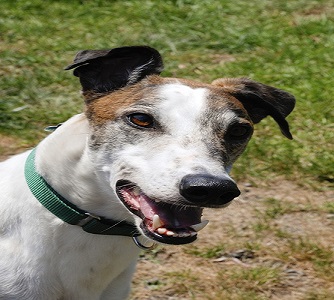}
            \end{subfigure}\vspace{.1ex}

\begin{subfigure}[t]{\textwidth}
                \includegraphics[width=\textwidth]{./img/Img/img_2008_000123.png}
            \end{subfigure}\vspace{.1ex}

\begin{subfigure}[t]{\textwidth}
                \includegraphics[width=\textwidth]{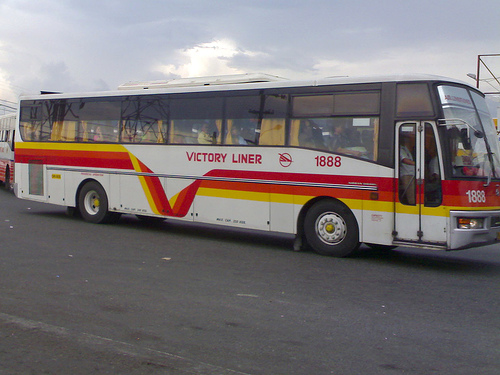}
            \end{subfigure}\vspace{.1ex}

\begin{subfigure}[t]{\textwidth}
                \includegraphics[width=\textwidth]{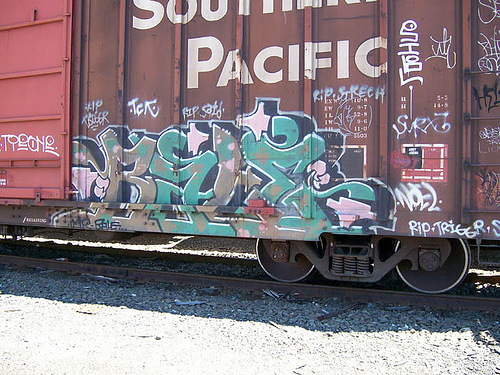}
            \end{subfigure}\vspace{.1ex}

            \begin{subfigure}[t]{\textwidth}
                \includegraphics[width=\textwidth]{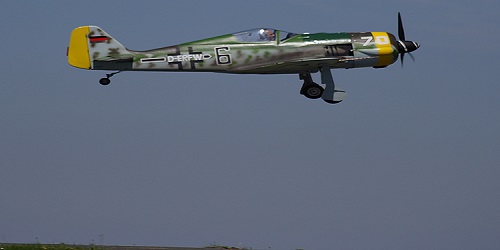}
            \subcaption{Image}
            \end{subfigure}
     \end{subfigure}
    \begin{subfigure}[t]{0.11\textwidth}
        \begin{subfigure}[t]{\textwidth}
                \includegraphics[width=\textwidth]{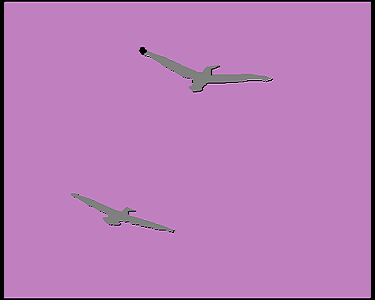}
            \end{subfigure}\vspace{.1ex}

        \begin{subfigure}[t]{\textwidth}
                \includegraphics[width=\textwidth]{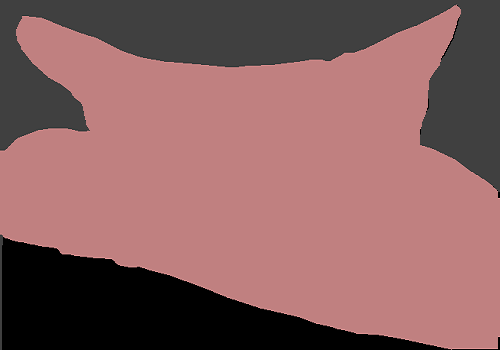}
            \end{subfigure}\vspace{.1ex}

        \begin{subfigure}[t]{\textwidth}
                \includegraphics[width=\textwidth]{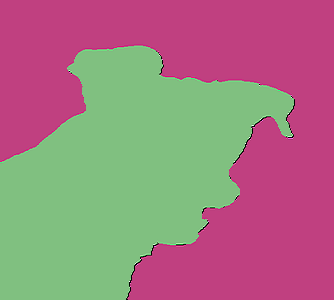}
            \end{subfigure}\vspace{.1ex}

 \begin{subfigure}[t]{\textwidth}
                \includegraphics[width=\textwidth]{./img/GT/gt_2008_000123.png}
            \end{subfigure}\vspace{.1ex}

\begin{subfigure}[t]{\textwidth}
                \includegraphics[width=\textwidth]{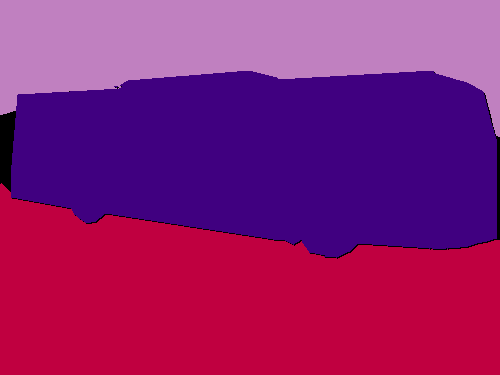}
            \end{subfigure}\vspace{.1ex}

\begin{subfigure}[t]{\textwidth}
                \includegraphics[width=\textwidth]{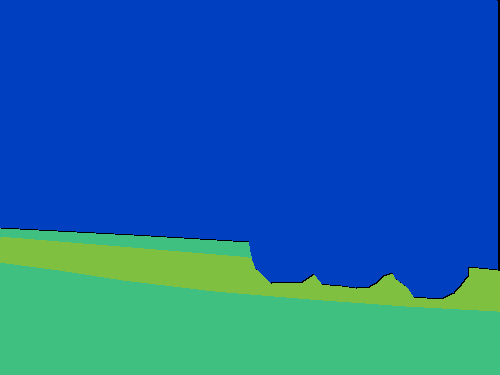}
            \end{subfigure}\vspace{.1ex}

            \begin{subfigure}[t]{\textwidth}
                \includegraphics[width=\textwidth]{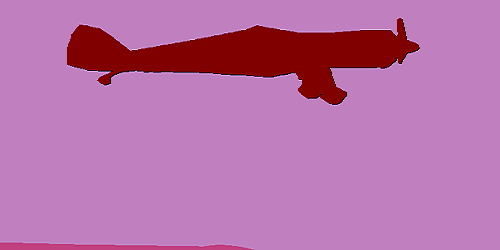}
     \captionsetup{justification=centering}       
	\subcaption{GT}
            \end{subfigure}
    \end{subfigure}
    \begin{subfigure}[t]{0.11\textwidth}
        \begin{subfigure}[t]{\textwidth}
                \includegraphics[width=\textwidth]{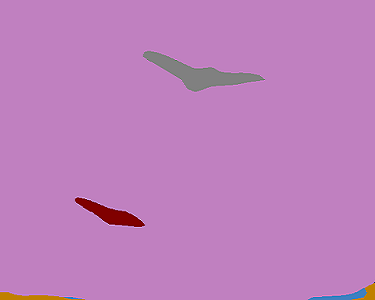}
            \end{subfigure}\vspace{.1ex}

        \begin{subfigure}[t]{\textwidth}
                \includegraphics[width=\textwidth]{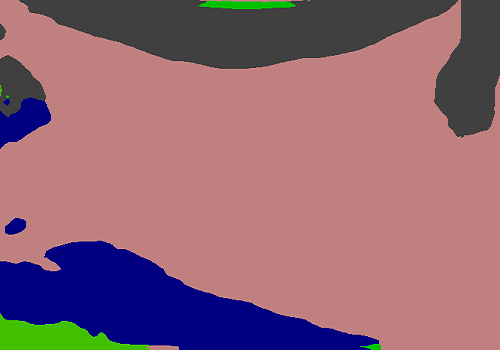}
            \end{subfigure}\vspace{.1ex}

        \begin{subfigure}[t]{\textwidth}
                \includegraphics[width=\textwidth]{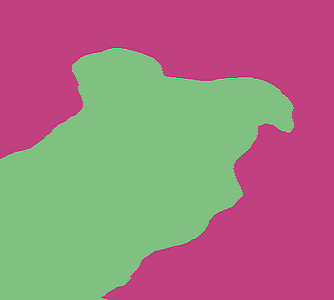}
            \end{subfigure}\vspace{.1ex}

\begin{subfigure}[t]{\textwidth}
                \includegraphics[width=\textwidth]{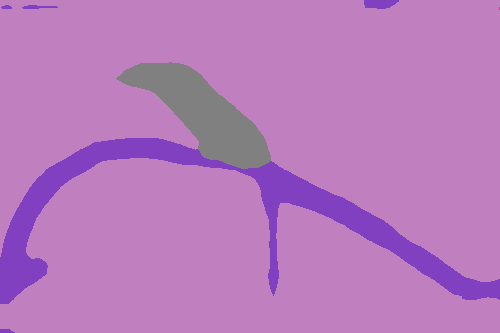}
            \end{subfigure}\vspace{.1ex}

\begin{subfigure}[t]{\textwidth}
                \includegraphics[width=\textwidth]{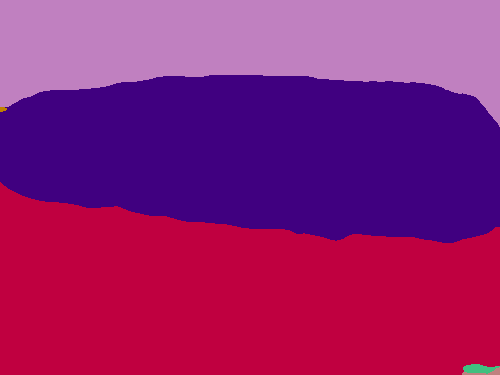}
            \end{subfigure}\vspace{.1ex}

\begin{subfigure}[t]{\textwidth}
                \includegraphics[width=\textwidth]{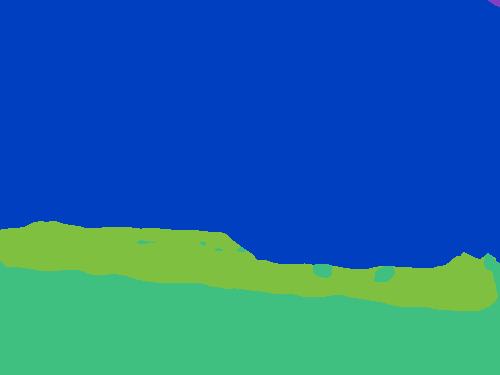}
            \end{subfigure}\vspace{.1ex}

            \begin{subfigure}[t]{\textwidth}
                \includegraphics[width=\textwidth]{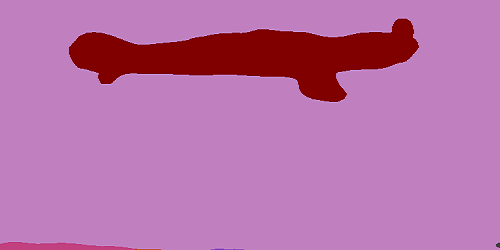}
            \captionsetup{justification=centering}
            \subcaption{Before LaU}
        \label{6c}
            \end{subfigure}
    \end{subfigure}
    \begin{subfigure}[t]{0.11\textwidth}
        \begin{subfigure}[t]{\textwidth}
                \includegraphics[width=\textwidth]{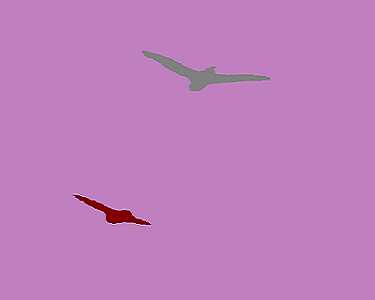}
            \end{subfigure}\vspace{.1ex}

        \begin{subfigure}[t]{\textwidth}
                \includegraphics[width=\textwidth]{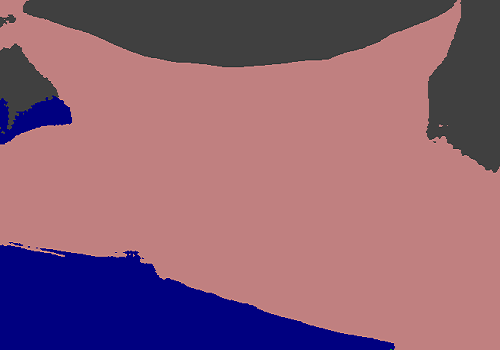}
            \end{subfigure}\vspace{.1ex}

        \begin{subfigure}[t]{\textwidth}
                \includegraphics[width=\textwidth]{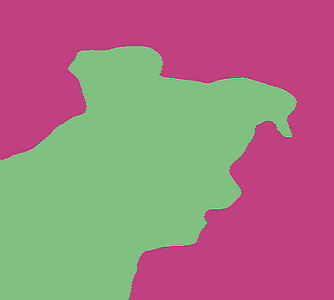}
            \end{subfigure}\vspace{.1ex}

\begin{subfigure}[t]{\textwidth}
                \includegraphics[width=\textwidth]{./img/results_w_offsets/2008_000123.png}
            \end{subfigure}\vspace{.1ex}

\begin{subfigure}[t]{\textwidth}
                \includegraphics[width=\textwidth]{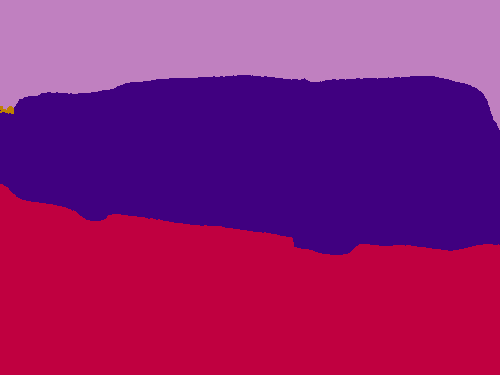}
            \end{subfigure}\vspace{.1ex}

\begin{subfigure}[t]{\textwidth}
                \includegraphics[width=\textwidth]{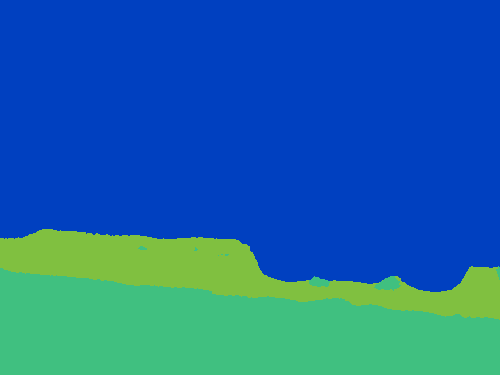}
            \end{subfigure}\vspace{.1ex}

            \begin{subfigure}[t]{\textwidth}
                \includegraphics[width=\textwidth]{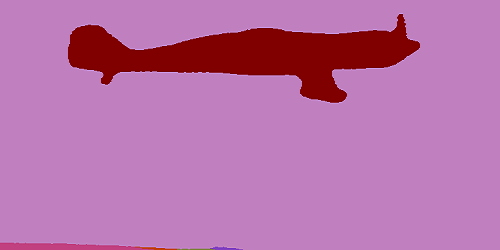}
            \subcaption{After LaU}
        \label{6d}
            \end{subfigure}
    \end{subfigure}
        \caption{Illustrations of the effect of predicted offsets generated by LaU$_{\text{off}_{4\times}}$. We visualize the segmentation results of EncNet+LaU before and after LaU.}
\label{view2}
\end{figure}

We show the PASCAL-Context visual results in Figure \ref{view}. LaU labels out the missing parts in EncNet and corrects the misclassifications. More examples in ADE20K and the visual comparison between offset-guided loss and regression loss are shown in the appendix (Figure \ref{cmp2}). To further illustrate the effect of location-aware mechanism, we compare the segmentation results before and after LaU in Figure \ref{view2}. LaU first produces coarse segmentation of objects with rough boundary (in Figure \ref{6c}), then refines the boundary using offsets (in Figure \ref{6d}). The whole segmentation process of LaU is from coarse to fine.

\begin{table}
  \caption{ADE20K \textit{val} set results with top-performing models (using mutli-scale testing) .}
  \label{sample-table}
  \centering
 \begin{tabular}{llcc}
\hline    
Method    & Backbone & pixAcc\%  & mIoU\%\\
\hline  \hline  
EncNet \cite{EncNet} & ResNet-50 & 79.73 & 41.11 \\
EncNet+LaU$_{\text{off}_{4\times}}$ & ResNet-50 & \textbf{80.53} & \textbf{42.78} \\
EncNet+LaU$_{\text{reg}_{4\times}}$ & ResNet-50 & 80.28 & 42.41 \\
\hline
HRNet \cite{HRR} & HRNetV2-W48  & -- & 43.20\\
PSPNet \cite{PSPNet} & ResNet-152 & 81.38 & 43.51\\
EncNet \cite{EncNet} & ResNet-101 & 81.69 & 44.65\\
PSPNet \cite{PSPNet} & ResNet-269 & \textbf{81.69} & 44.94\\
EncNet+LaU$_{\text{off}_{4\times}}$ & ResNet-101 & 81.21 & 44.55 \\
EncNet+LaU$_{\text{reg}_{4\times}}$ & ResNet-101 & 81.49  & \textbf{45.02} \\
\hline  
  \end{tabular}
\label{ade20k_val}
\end{table}

\begin{table}
  \caption{Results on ADE20K \textit{test} set. Best entries in COCO-Place challenge 2017 are listed.}
  \label{sample-table}
  \centering
 \begin{tabular}{llcc}
\hline    
 Team & Model  & Final Score $\uparrow$ & mIoU\%\\
\hline  \hline  
 PSPNet \cite{PSPNet} & ResNet-269 & 0.5538 & --\\
 WinterISComing$^{2_{\text{nd}}}$ & -- & 0.5544 & --\\
CASIA\_IVA\_JD$^{1_{\text{st}}}$ & -- & 0.5547 & --\\
 EncNet \cite{EncNet} & ResNet-101 & 0.5567 & --\\
 JPU \cite{JPU} & ResNet-101 & 0.5584 & 38.39\\
 EncNet+LaU$_{\text{off}_{4\times}}$ & ResNet-101 & \textbf{0.5641} & 39.04\\
 EncNet+LaU$_{\text{reg}_{4\times}}$ & ResNet-101 & 0.5632 & \textbf{39.14}\\
\hline  
  \end{tabular}
\label{ade20k_test}
\end{table}

\subsection{Applying to Decoder}
Recent state-of-the-arts \cite{ASPP, deeplabv3} utilize upsampling modules in decoders to combine the high-resolution features with low-resolution ones. To show the general applicability of proposed method, we replace the bilinear upsampling in DeepLabv3$+$ \cite{deeplabv3} with LaU. The performance of ResNet-50+LaU ($+2.03\%$ mIoU in Table \ref{deeplab}) has approached the original ResNet-101 backbone results. Even with strong ResNet-101 baseline, LaU can still boost the performance by one point.

\begin{table}
  \caption{mIoU over Pascal VOC 2012 \textit{val} set using DeepLabv3$+$ \cite{deeplabv3}.}
  \label{sample-table}
  \centering
 \begin{tabular}{llcc}
\hline    
Method    & Upsampling & Backbone   & mIoU\%\\
\hline  \hline  
    \multirow{4}{*}{DeepLabv3$+$}  & Bilinear & ResNet-50  & 76.52 \\
    & LaU$_{4\times}$ & ResNet-50 & \textbf{78.55} \\
    & Bilinear & ResNet-101  & 78.85 \\ 
   & LaU$_{4\times}$ & ResNet-101 & \textbf{80.16} \\ 
\hline
  \end{tabular}
\label{deeplab}
\end{table}

\section{Conclusion}
In this paper, we provide a new perspective for optimizing the semantic segmentation problem. We decompose the original pixel-level classification problem into offsets prediction and classification, which introduces the idea of location prediction into semantic segmentation. Based on this fresh viewpoint, we propose the location-aware upsampling and location-aware losses. Our models achieve promising performance compared with various baseline methods. The effectiveness of learning offsets is also verified through qualitative and quantitative results.

{\small
\bibliographystyle{ieee_fullname}
\bibliography{egbib}
}

\appendix

\section{Run-time Speed}
In this section, we employ the frame per second (FPS) as the evaluation metirc to show the time efficiency about the proposed method. Compared with the traditional bilinear upsampling, LaU consistently improves the performance at the cost of negligible extra time complexity. We report the FPS averaged over 100 iterations.  As shown in Table \ref{fps}, LaU is competitive with the bilinear upsamling network at actual run-time speed.

\begin{table}[th]
  \caption{Actual inference time on a GTX 1080Ti GPU with $512\times512$ input size. ``OS" refers to the output stride. }
  \label{sample-table}
  \centering
\begin{tabular}{l | cccc}
\hline 
    Head     &   OS   & Backbone &  Upsampling & FPS\\
\hline   
\hline
\multirow{8}{*}{EncNet \cite{EncNet}} & 8 & ResNet-50 & Bilinear & 18.60\\
 & 8 & ResNet-50 & LaU$_{4\times}$ & 18.39\\
\cline{2-5}
& 32 & ResNet-50 & Bilinear & 70.80\\
& 32 & ResNet-50 & LaU$_{4\times}$ & 69.84\\
\cline{2-5}
& 8 & ResNet-101 & Bilinear & 11.66\\
 & 8 & ResNet-101 & LaU$_{4\times}$ & 11.49\\
\cline{2-5}
& 32 & ResNet-101 & Bilinear & 47.37\\
& 32 & ResNet-101 & LaU$_{4\times}$ & 46.46\\
\hline 
\hline 
\multirow{4}{*}{ASPP \cite{ASPP}} & 8 & ResNet-50 & Bilinear & 15.93\\
 & 8 & ResNet-50 & LaU$_{4\times}$ & 15.72\\
\cline{2-5}
& 8 & ResNet-101 & Bilinear & 10.48\\
& 8 & ResNet-101 & LaU$_{4\times}$ & 10.32\\
\hline 
\hline 
  \end{tabular}
\label{fps}
\end{table}

\section{Implementation Details}
Our experiments base on the PyTorch framework \cite{paszke2017automatic} and pre-trained models provided by EncNet \cite{EncNet}. We use the official open source code \cite{ZhuHLD19} to conduct the deformable convolution experiments. In this section, we employs ResNet-50 as the backbone without multi-scale testing. The results are reported on PASCAL Context \textit{val} set. To make a fair comparison, we follow the same setting in our main paper.

\begin{figure}[t]
\center
    \begin{subfigure}[t]{0.155\textwidth}
            \begin{subfigure}[t]{\textwidth}
                \includegraphics[width=\textwidth]{./img/Img/img_2009_002618.jpg}
            \end{subfigure}\vspace{.1ex}
            \begin{subfigure}[t]{\textwidth}
                \includegraphics[width=\textwidth]{./img/Img/img_2008_000123.png}
            \end{subfigure}\vspace{.1ex}
         \begin{subfigure}[t]{\textwidth}
                \includegraphics[width=\textwidth]{./img/Img/img_2010_003270.jpg}
            \end{subfigure}\vspace{.1ex}
            \begin{subfigure}[t]{\textwidth}
                \includegraphics[width=\textwidth]{./img/Img/img_2009_004366.jpg}
            \subcaption{Image}
            \end{subfigure}
     \end{subfigure}
    \begin{subfigure}[t]{0.155\textwidth}
        \begin{subfigure}[t]{\textwidth}
                \includegraphics[width=\textwidth]{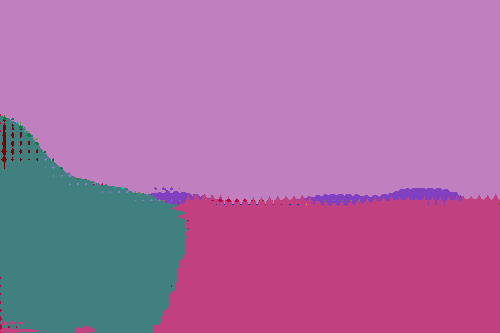}
            \end{subfigure}\vspace{.1ex}
        \begin{subfigure}[t]{\textwidth}
                \includegraphics[width=\textwidth]{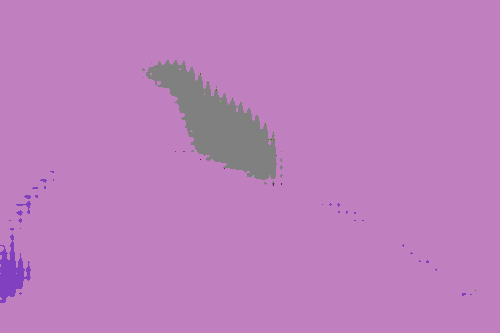}
            \end{subfigure}\vspace{.1ex}
        \begin{subfigure}[t]{\textwidth}
                \includegraphics[width=\textwidth]{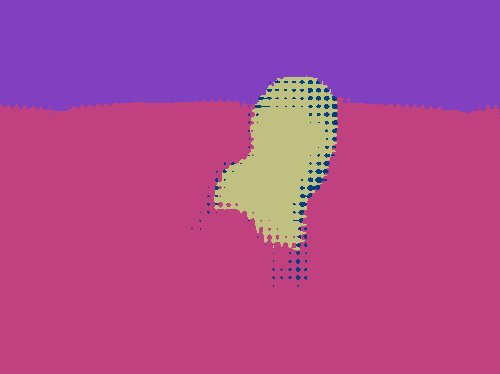}
            \end{subfigure}\vspace{.1ex}
            \begin{subfigure}[t]{\textwidth}
                \includegraphics[width=\textwidth]{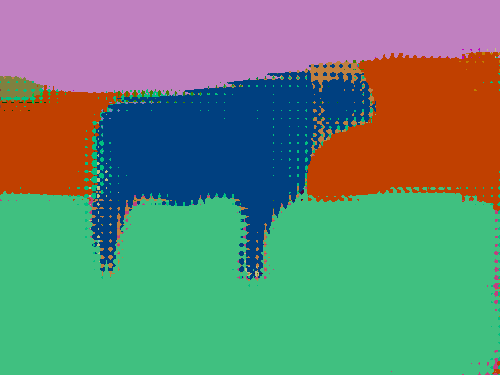}
            \captionsetup{justification=centering}
            \subcaption{ASPP Head}
        \label{d}
            \end{subfigure}
    \end{subfigure}
    \begin{subfigure}[t]{0.155\textwidth}
        \begin{subfigure}[t]{\textwidth}
                \includegraphics[width=\textwidth]{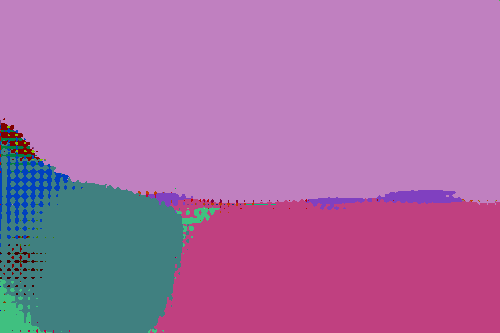}
            \end{subfigure}\vspace{.1ex}

        \begin{subfigure}[t]{\textwidth}
                \includegraphics[width=\textwidth]{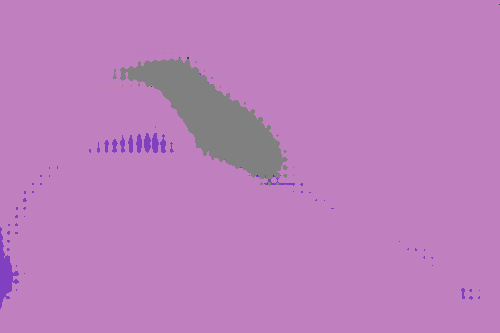}
            \end{subfigure}\vspace{.1ex}

        \begin{subfigure}[t]{\textwidth}
                \includegraphics[width=\textwidth]{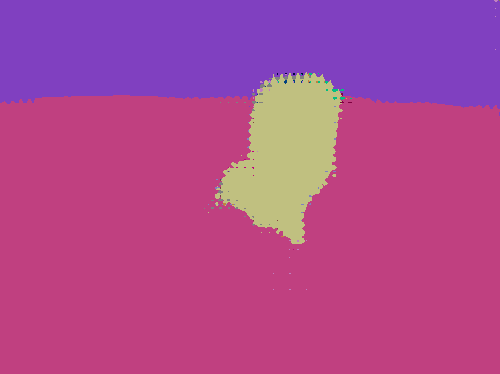}
            \end{subfigure}\vspace{.1ex}

            \begin{subfigure}[t]{\textwidth}
                \includegraphics[width=\textwidth]{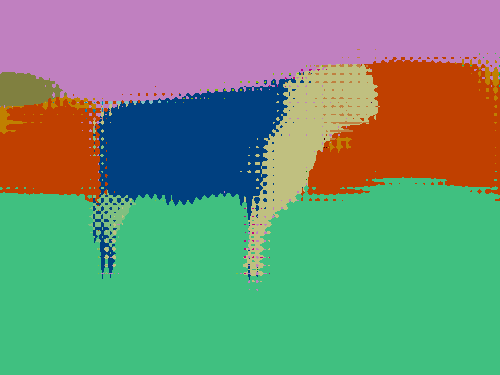}
	 \captionsetup{justification=centering}            
\subcaption{EncNet Head}
        \label{e}
            \end{subfigure}
    \end{subfigure}
        \caption{PixelShuffle$_{2\times}$ segmentation results. Figure (b) and (c) show the checkerboard artifacts. Although the last convolution layer (i.e., $59C_3$) may smooth the outputs, the results still consist of strange checkerboard pattern of artifacts.}
\label{pixleshuffle}
\end{figure}

\subsection{PixelShuffle}
We assume that the class number is 59. Following the popular setting in super-resolutions \cite{Lim_2017_CVPR_Workshops, ShiCHTABRW16, HeMWLY019}, the architecture of PixelShuffle-based upsamplings are as follow :
\begin{align*}
2_\times\text{: }&236C_3+\text{PixelShuffle}_{2\times}+59C_3\\
3_\times\text{: }&531C_3+\text{PixelShuffle}_{3\times}+59C_3\\
4_\times\text{: }&236C_3+\text{PixelShuffle}_{2\times}+236C_3+\text{PixelShuffle}_{2\times}+59C_3
\end{align*}
where $236C_3$ refers to the $3\times3$ convolution layer with 236 filters.

The intuitive idea is that, in the worst case, pixelshuffle should degrade into nearest neighbor upsampling (same feature map and weight for each shuffle group). However, the experiment results in Table 3a demostrates that PixelShuffle damages the performance. In this subsection, we show the segmentation results to reveal checkboard problems in PixelShuffle. As illustrated in Figure \ref{pixleshuffle}, both EncNet \cite{EncNet} head and ASPP \cite{ASPP} head suffer from the checkboard artifacts.

\begin{figure}[t]
\center
    \begin{subfigure}[t]{0.15\textwidth}
            \begin{subfigure}[t]{\textwidth}
                \includegraphics[width=\textwidth]{./img/Img/img_2009_002618.jpg}
            \end{subfigure}\vspace{.1ex}

            \begin{subfigure}[t]{\textwidth}
                \includegraphics[width=\textwidth]{./img/Img/img_2008_000123.png}
            \end{subfigure}\vspace{.1ex}

         \begin{subfigure}[t]{\textwidth}
                \includegraphics[width=\textwidth]{./img/Img/img_2010_003270.jpg}
            \end{subfigure}\vspace{.1ex}

            \begin{subfigure}[t]{\textwidth}
                \includegraphics[width=\textwidth]{./img/Img/img_2009_004366.jpg}
            \subcaption{Image}
            \end{subfigure}
     \end{subfigure}
    \begin{subfigure}[t]{0.15\textwidth}
        \begin{subfigure}[t]{\textwidth}
                \includegraphics[width=\textwidth]{./img/GT/gt_2009_002618.png}
            \end{subfigure}\vspace{.1ex}

        \begin{subfigure}[t]{\textwidth}
                \includegraphics[width=\textwidth]{./img/GT/gt_2008_000123.png}
            \end{subfigure}\vspace{.1ex}

        \begin{subfigure}[t]{\textwidth}
                \includegraphics[width=\textwidth]{./img/GT/gt_2010_003270.png}
            \end{subfigure}\vspace{.1ex}

            \begin{subfigure}[t]{\textwidth}
                \includegraphics[width=\textwidth]{./img/GT/gt_2009_004366.png}
     \captionsetup{justification=centering}       
	\subcaption{GT}
            \end{subfigure}
    \end{subfigure}
    \begin{subfigure}[t]{0.15\textwidth}
        \begin{subfigure}[t]{\textwidth}
                \includegraphics[width=\textwidth]{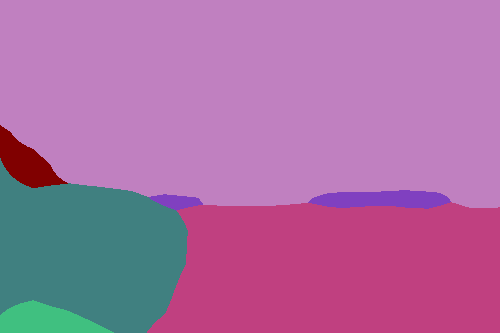}
            \end{subfigure}\vspace{.1ex}

        \begin{subfigure}[t]{\textwidth}
                \includegraphics[width=\textwidth]{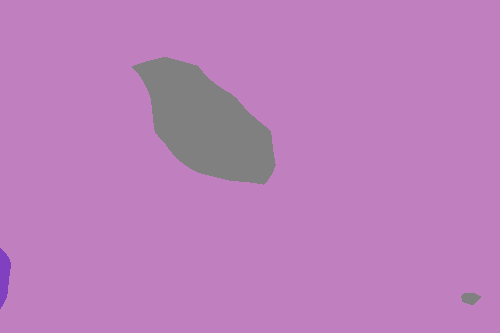}
            \end{subfigure}\vspace{.1ex}

        \begin{subfigure}[t]{\textwidth}
                \includegraphics[width=\textwidth]{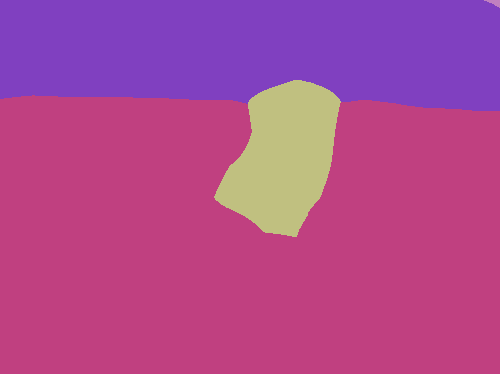}
            \end{subfigure}\vspace{.1ex}

            \begin{subfigure}[t]{\textwidth}
                \includegraphics[width=\textwidth]{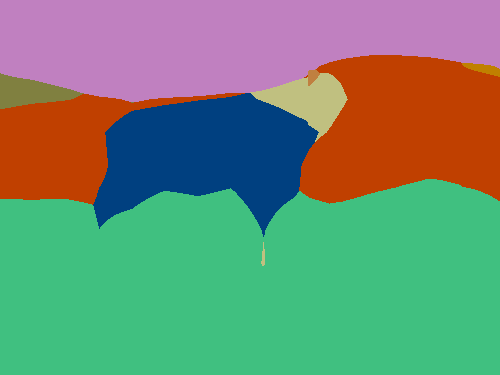}
 \captionsetup{justification=centering}            
\subcaption{Deformable}
        \label{e}
            \end{subfigure}
    \end{subfigure}
\caption{Visualization results using deformable bilinear upsampling with EncNet-ResNet50 \cite{EncNet}.}
\label{deformable}
\end{figure}

\subsection{Deformable Convolution}
We replace the traditional bilinear upsamling in encoder-decoder networks with ``Deformable $3\times3$ Convolution \cite{DaiQXLZHW17} + Blinear$_{8\times}$" module. Since ``Blinear + Deformable  Convolution" will hughly increase the extra computing cost, we only conduct experiments under the first setting.

Although deformable convolution has been proved to improve the segmentation task when properly used in feature extraction, we show that the ``deformable upsampling" is probably not appropriate for the final prediction. As illustrated in Figure \ref{deformable}, deformable bilinear upsampling tends to focus on the main body of the objects yet miss the details, which consists with the design  of deformable convolution networks.

\section{Visualization Results}
\begin{figure}
\center
    \begin{subfigure}[t]{0.11\textwidth}
            \begin{subfigure}[t]{\textwidth}
                \includegraphics[width=\textwidth]{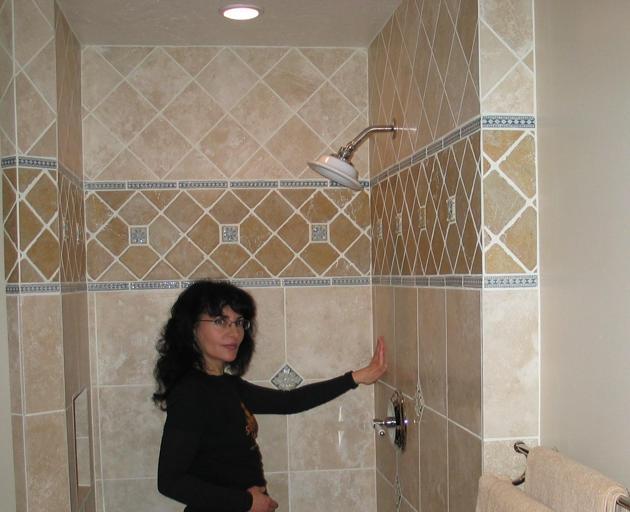}
            \end{subfigure}\vspace{.1ex}

            \begin{subfigure}[t]{\textwidth}
                \includegraphics[width=\textwidth]{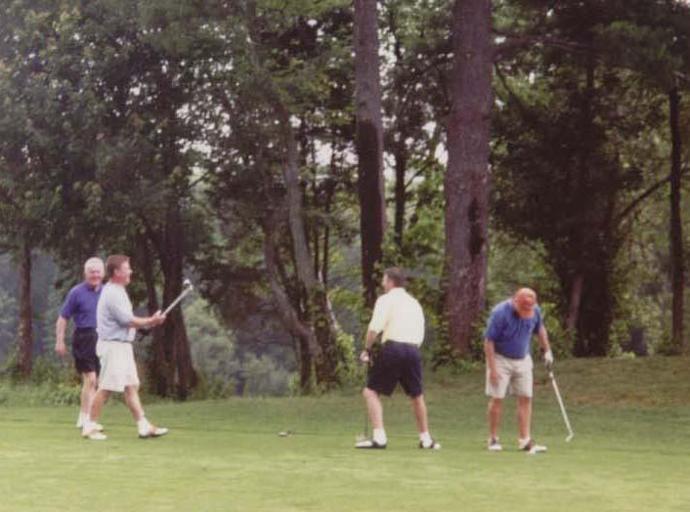}
            \end{subfigure}\vspace{.1ex}

         \begin{subfigure}[t]{\textwidth}
                \includegraphics[width=\textwidth]{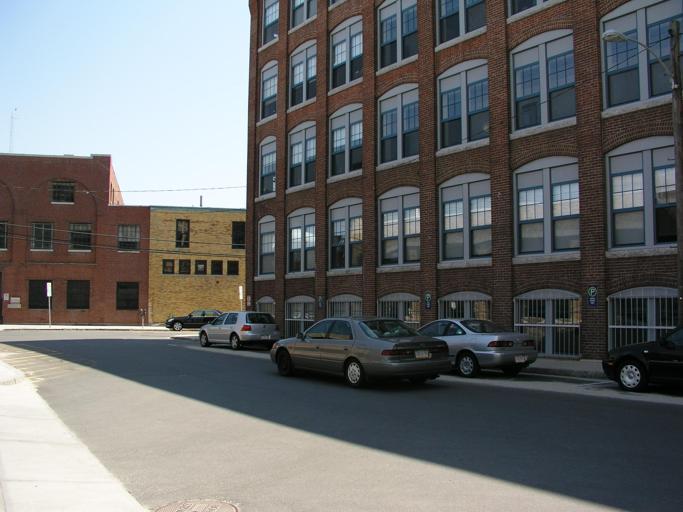}
            \end{subfigure}\vspace{.1ex}

            \begin{subfigure}[t]{\textwidth}
                \includegraphics[width=\textwidth]{./img/Img/img_ADE_val_00000767.jpg}
            \subcaption{Image}
            \end{subfigure}
     \end{subfigure}
    \begin{subfigure}[t]{0.11\textwidth}
        \begin{subfigure}[t]{\textwidth}
                \includegraphics[width=\textwidth]{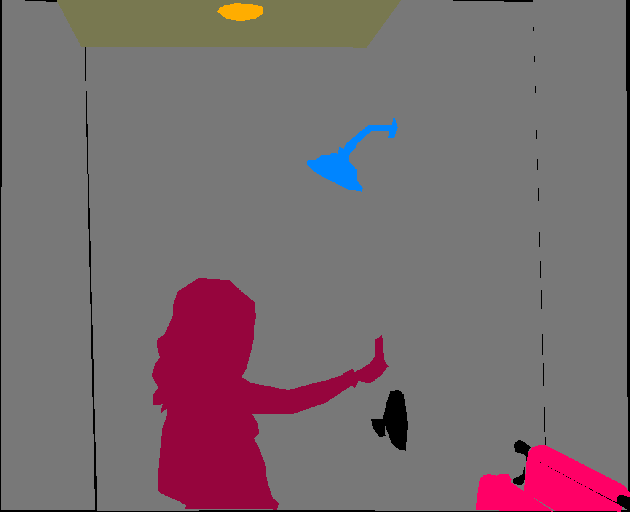}
            \end{subfigure}\vspace{.1ex}

        \begin{subfigure}[t]{\textwidth}
                \includegraphics[width=\textwidth]{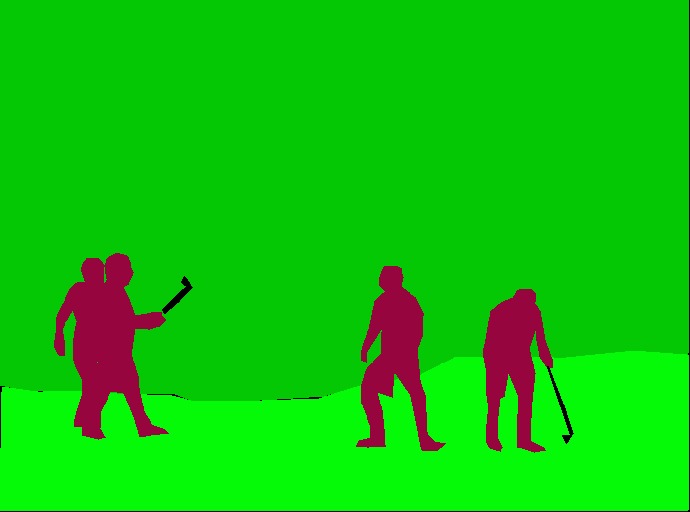}
            \end{subfigure}\vspace{.1ex}

        \begin{subfigure}[t]{\textwidth}
                \includegraphics[width=\textwidth]{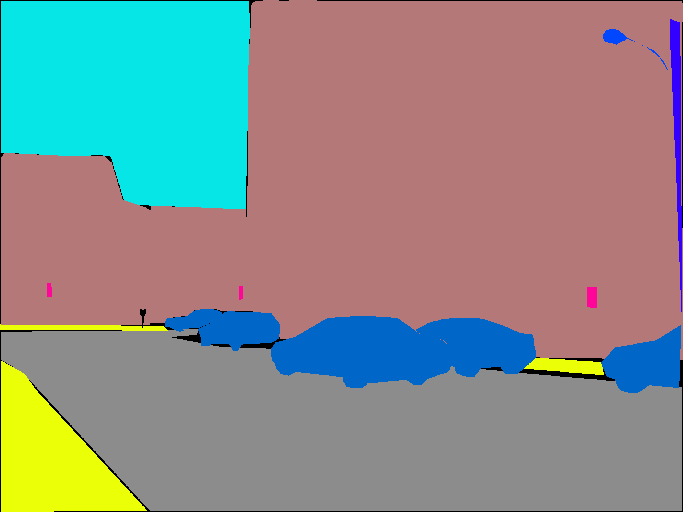}
            \end{subfigure}\vspace{.1ex}

            \begin{subfigure}[t]{\textwidth}
                \includegraphics[width=\textwidth]{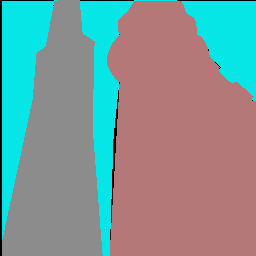}
     \captionsetup{justification=centering}       
	\subcaption{GT}
            \end{subfigure}
    \end{subfigure}
    \begin{subfigure}[t]{0.11\textwidth}
        \begin{subfigure}[t]{\textwidth}
                \includegraphics[width=\textwidth]{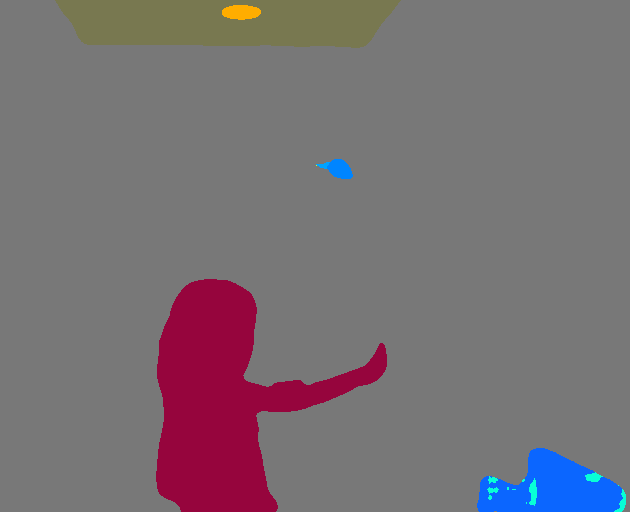}
            \end{subfigure}\vspace{.1ex}

        \begin{subfigure}[t]{\textwidth}
                \includegraphics[width=\textwidth]{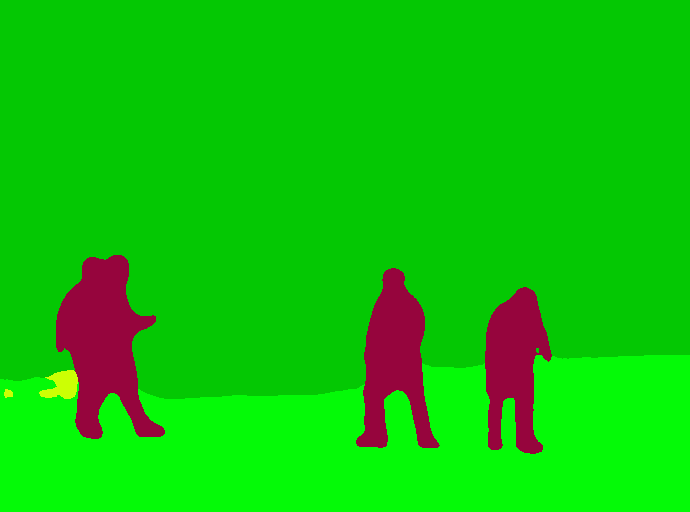}
            \end{subfigure}\vspace{.1ex}

        \begin{subfigure}[t]{\textwidth}
                \includegraphics[width=\textwidth]{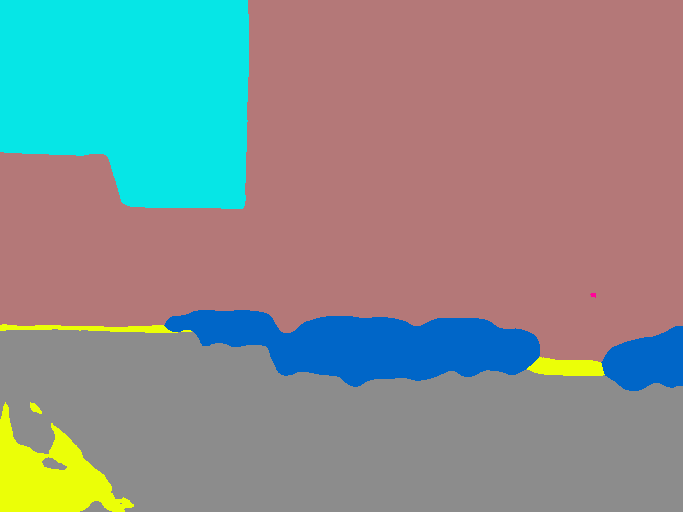}
            \end{subfigure}\vspace{.1ex}

            \begin{subfigure}[t]{\textwidth}
                \includegraphics[width=\textwidth]{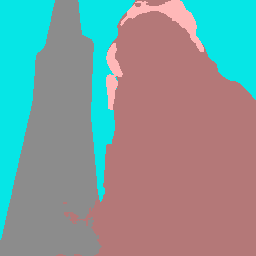}
            \captionsetup{justification=centering}
            \subcaption{EncNet}
        \label{d}
            \end{subfigure}
    \end{subfigure}
    \begin{subfigure}[t]{0.11\textwidth}
        \begin{subfigure}[t]{\textwidth}
                \includegraphics[width=\textwidth]{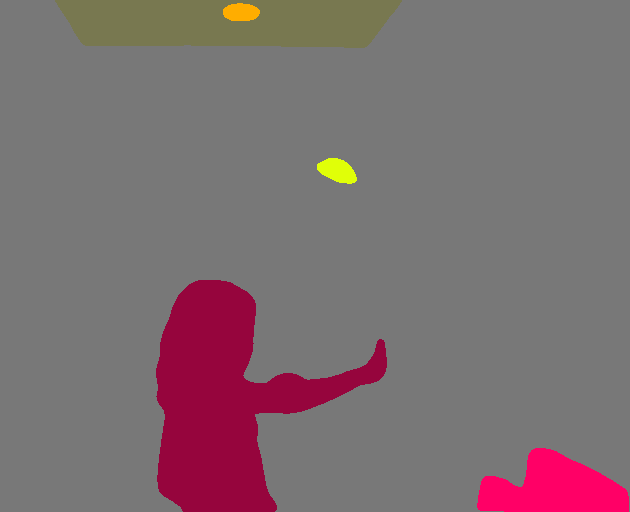}
            \end{subfigure}\vspace{.1ex}

        \begin{subfigure}[t]{\textwidth}
                \includegraphics[width=\textwidth]{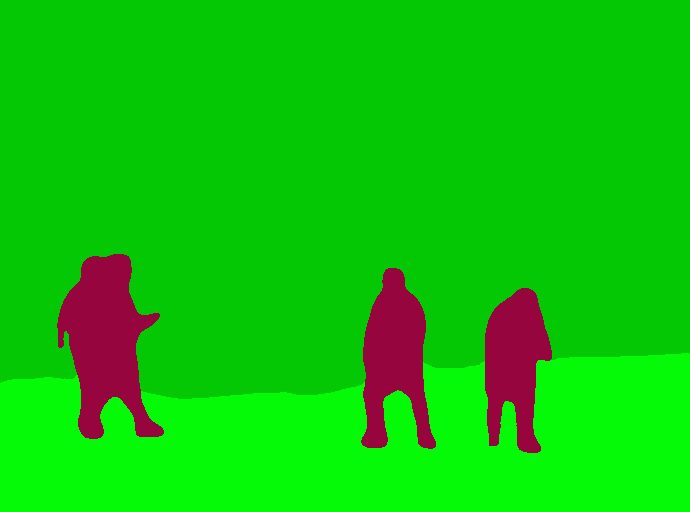}
            \end{subfigure}\vspace{.1ex}

        \begin{subfigure}[t]{\textwidth}
                \includegraphics[width=\textwidth]{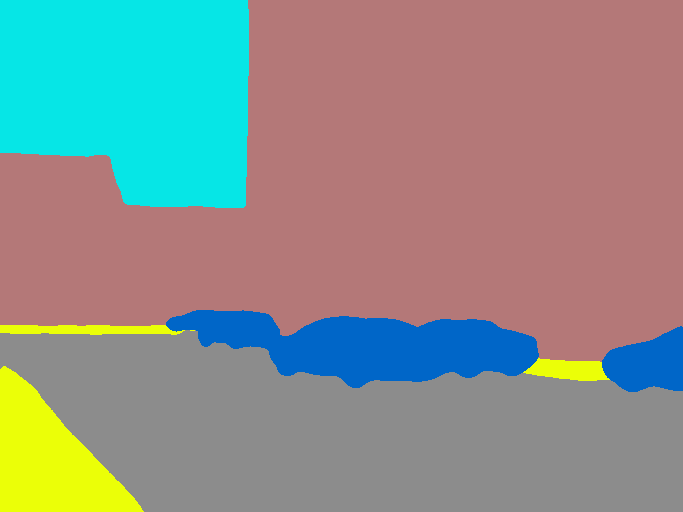}
            \end{subfigure}\vspace{.1ex}

            \begin{subfigure}[t]{\textwidth}
                \includegraphics[width=\textwidth]{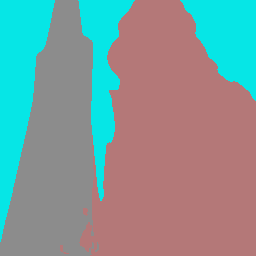}
            \subcaption{LaU}
        \label{e}
            \end{subfigure}
    \end{subfigure}
        \caption{Visual improvements on ADE20K with ResNet101 backbone. ``EncNet" is the standard bilinear upsampling. ``LaU" refers to LaU$_{\text{off}_{4\times}}$.}
\label{cmp}
\end{figure}

\begin{figure}
\center
    \begin{subfigure}[t]{0.11\textwidth}
            \begin{subfigure}[t]{\textwidth}
                \includegraphics[width=\textwidth]{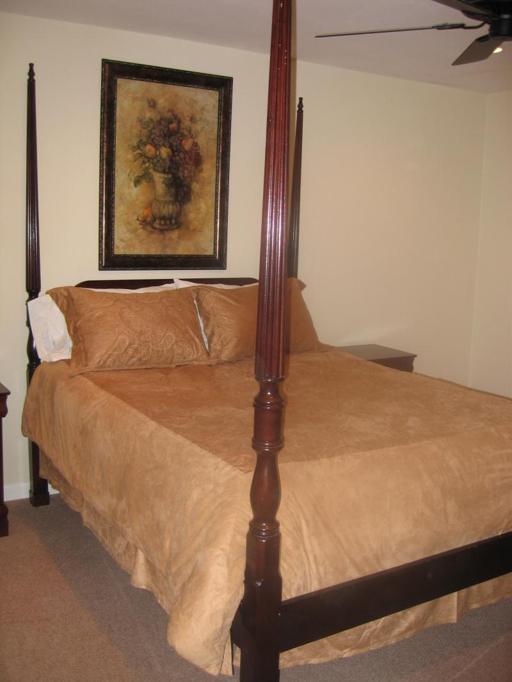}
            \end{subfigure}\vspace{.1ex}

            \begin{subfigure}[t]{\textwidth}
                \includegraphics[width=\textwidth]{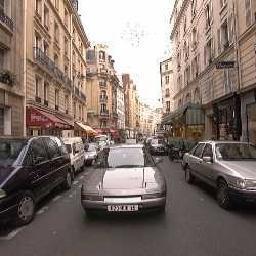}
            \end{subfigure}\vspace{.1ex}

         \begin{subfigure}[t]{\textwidth}
                \includegraphics[width=\textwidth]{./img/Img/img_ADE_val_00000925.jpg}
            \end{subfigure}\vspace{.1ex}

\begin{subfigure}[t]{\textwidth}
                \includegraphics[width=\textwidth]{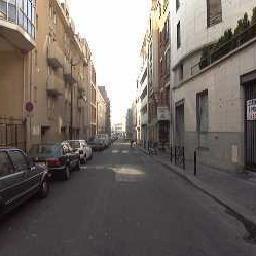}
            \end{subfigure}\vspace{.1ex}

            \begin{subfigure}[t]{\textwidth}
                \includegraphics[width=\textwidth]{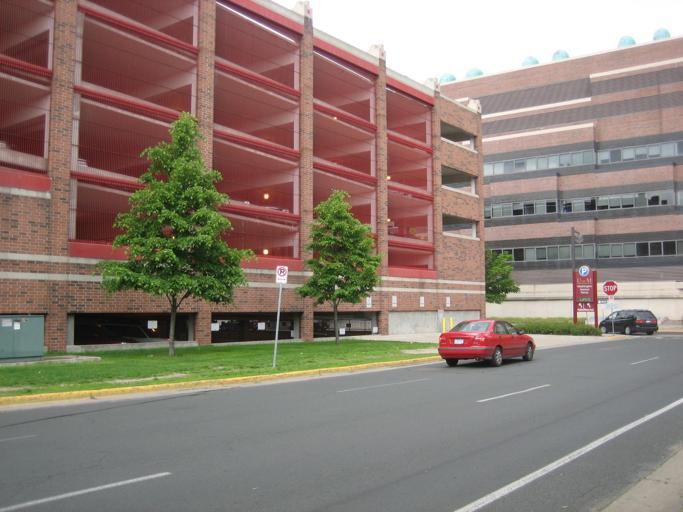}
            \subcaption{Image}
            \end{subfigure}
     \end{subfigure}
    \begin{subfigure}[t]{0.11\textwidth}
        \begin{subfigure}[t]{\textwidth}
                \includegraphics[width=\textwidth]{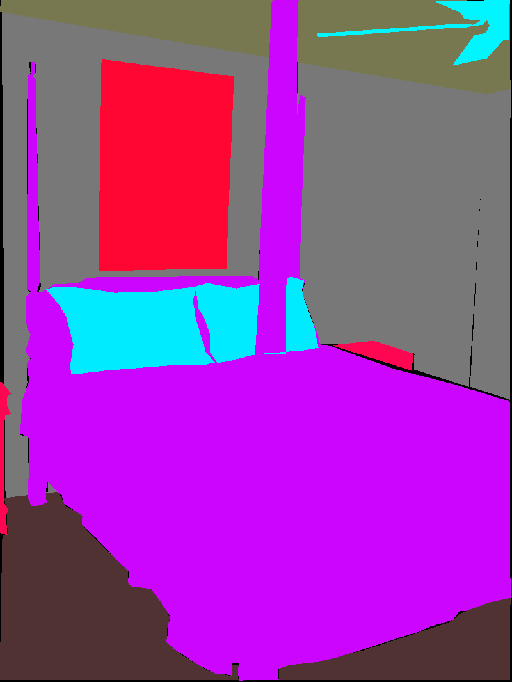}
            \end{subfigure}\vspace{.1ex}

        \begin{subfigure}[t]{\textwidth}
                \includegraphics[width=\textwidth]{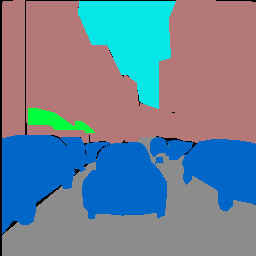}
            \end{subfigure}\vspace{.1ex}

        \begin{subfigure}[t]{\textwidth}
                \includegraphics[width=\textwidth]{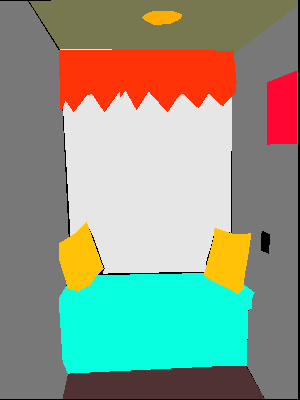}
            \end{subfigure}\vspace{.1ex}

\begin{subfigure}[t]{\textwidth}
                \includegraphics[width=\textwidth]{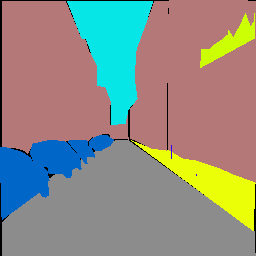}
            \end{subfigure}\vspace{.1ex}

            \begin{subfigure}[t]{\textwidth}
                \includegraphics[width=\textwidth]{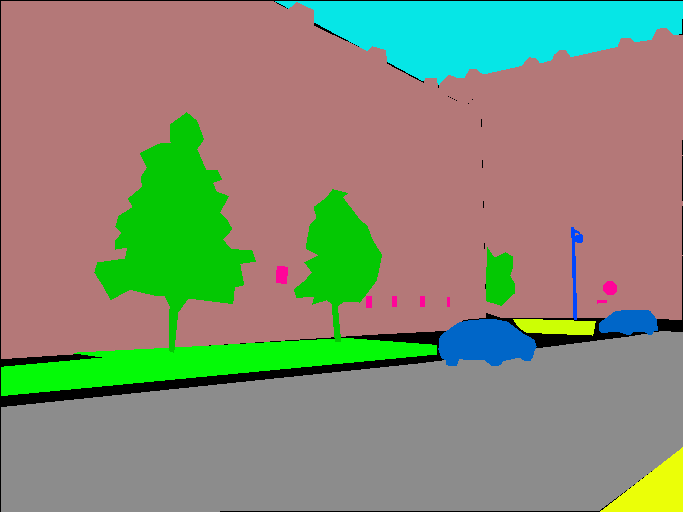}
     \captionsetup{justification=centering}       
	\subcaption{GT}
            \end{subfigure}
    \end{subfigure}
    \begin{subfigure}[t]{0.11\textwidth}
        \begin{subfigure}[t]{\textwidth}
                \includegraphics[width=\textwidth]{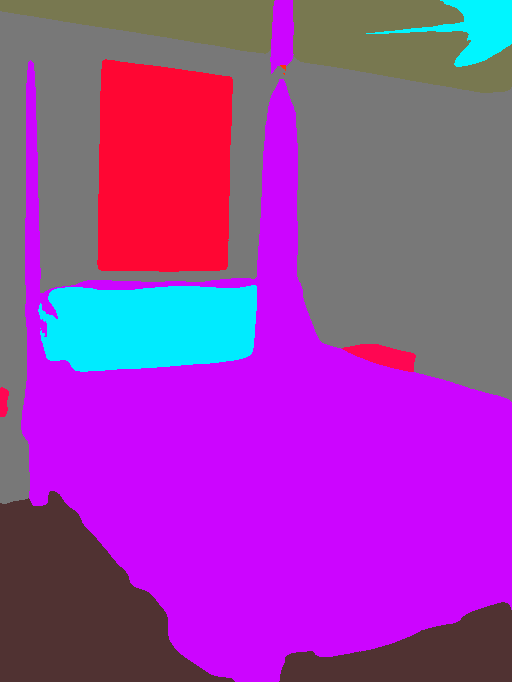}
            \end{subfigure}\vspace{.1ex}

        \begin{subfigure}[t]{\textwidth}
                \includegraphics[width=\textwidth]{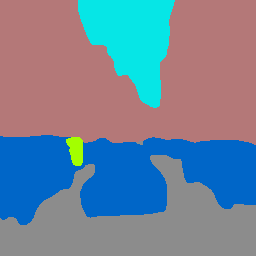}
            \end{subfigure}\vspace{.1ex}

        \begin{subfigure}[t]{\textwidth}
                \includegraphics[width=\textwidth]{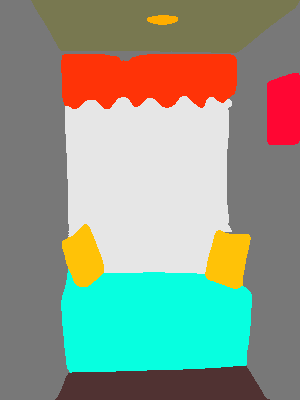}
            \end{subfigure}\vspace{.1ex}

\begin{subfigure}[t]{\textwidth}
                \includegraphics[width=\textwidth]{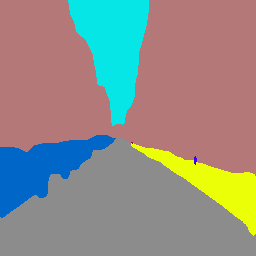}
            \end{subfigure}\vspace{.1ex}

            \begin{subfigure}[t]{\textwidth}
                \includegraphics[width=\textwidth]{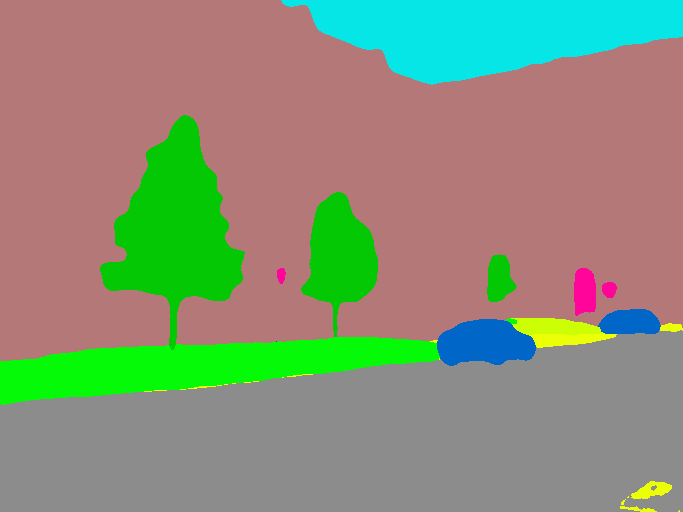}
            \captionsetup{justification=centering}
            \subcaption{offset-\\ guided loss}
        \label{d}
            \end{subfigure}
    \end{subfigure}
    \begin{subfigure}[t]{0.11\textwidth}
        \begin{subfigure}[t]{\textwidth}
                \includegraphics[width=\textwidth]{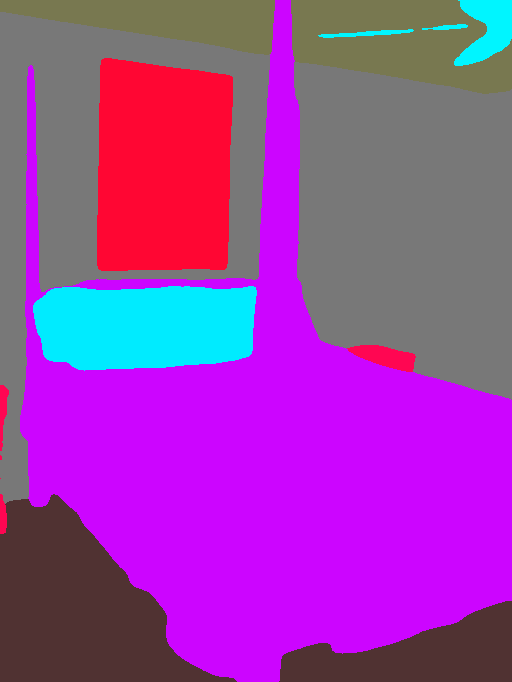}
            \end{subfigure}\vspace{.1ex}

        \begin{subfigure}[t]{\textwidth}
                \includegraphics[width=\textwidth]{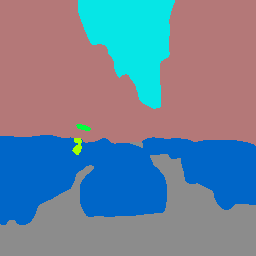}
            \end{subfigure}\vspace{.1ex}

        \begin{subfigure}[t]{\textwidth}
                \includegraphics[width=\textwidth]{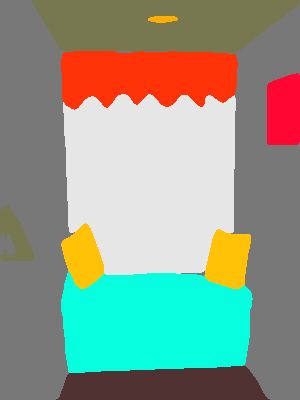}
            \end{subfigure}\vspace{.1ex}

\begin{subfigure}[t]{\textwidth}
                \includegraphics[width=\textwidth]{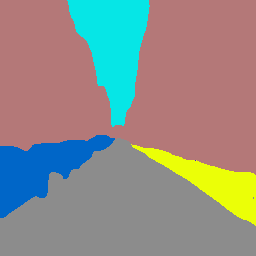}
            \end{subfigure}\vspace{.1ex}

            \begin{subfigure}[t]{\textwidth}
                \includegraphics[width=\textwidth]{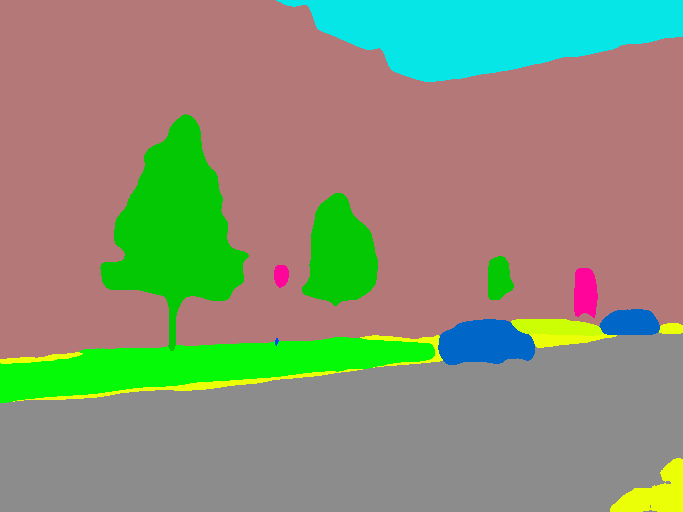}
 \captionsetup{justification=centering}            
\subcaption{regression\\loss}
        \label{e}
            \end{subfigure}
    \end{subfigure}
        \caption{Visual comparison on ADE20K \textit{val} set.}
\label{cmp2}
\end{figure}

In this section, we show the visual results from ADE20K dataset. For the qualitative analysis, we compare the outputs generated by LaU and bilinear upsampling, shown in Figure \ref{cmp}. Since offset-guided loss and regression loss achieve similar mIoU and pixel accuracy, we visualize the segmentation results on \textit{val} set in Figure \ref{cmp2} and make a qualitative evaluation. Regression loss performs slightly better than offset-guided loss, which is consistent with the mIoU results.

\end{document}